\documentclass[runningheads]{llncs}
\usepackage{graphicx}

\usepackage{tikz}
\usepackage{comment}
\usepackage{amsmath,amssymb} 
\usepackage{color}
\usepackage{bbold}
\usepackage{amssymb}
\usepackage{booktabs}
\usepackage{multirow}
\usepackage{makecell}
\usepackage{wrapfig}
\usepackage[pagebackref,breaklinks,colorlinks]{hyperref}
\usepackage[accsupp]{axessibility}  

\usepackage[capitalize]{cleveref}
\crefname{section}{Sec.}{Secs.}
\Crefname{section}{Section}{Sections}
\Crefname{table}{Table}{Tables}
\crefname{table}{Tab.}{Tabs.}

\begin{document}
\pagestyle{headings}
\mainmatter
\def\ECCVSubNumber{6356}  

\title{FCAF3D: Fully Convolutional Anchor-Free 3D Object Detection} 


\titlerunning{FCAF3D: Fully Convolutional Anchor-Free 3D Object Detection}

\author{Danila Rukhovich\and
Anna Vorontsova\and
Anton Konushin}

\authorrunning{D. Rukhovich et al.}
%
\institute{Samsung AI Center, Moscow\\
\email{\{d.rukhovich, a.vorontsova, a.konushin\}@samsung.com}}

\maketitle

\begin{abstract}
Recently, promising applications in robotics and augmented reality have attracted considerable attention to 3D object detection from point clouds. In this paper, we present FCAF3D --- a first-in-class fully convolutional anchor-free indoor 3D object detection method. It is a simple yet effective method that uses a voxel representation of a point cloud and processes voxels with sparse convolutions. FCAF3D can handle large-scale scenes with minimal runtime through a single fully convolutional feed-forward pass. Existing 3D object detection methods make prior assumptions on the geometry of objects, and we argue that it limits their generalization ability. To eliminate prior assumptions, we propose a novel parametrization of oriented bounding boxes that allows obtaining better results in a purely data-driven way. The proposed method achieves state-of-the-art 3D object detection results in terms of mAP@0.5 on ScanNet V2 (\textbf{+4.5}), SUN RGB-D (\textbf{+3.5}), and S3DIS (\textbf{+20.5}) datasets. The code and models are available at \\\url{https://github.com/samsunglabs/fcaf3d}.

\end{abstract}

\section{Introduction}

3D object detection from point clouds aims at simultaneous localization and recognition of 3D objects given a 3D point set. As a core technique for 3D scene understanding, it is widely applied in autonomous driving, robotics, and AR.

\begin{figure}[h!]
    \centering
        \includegraphics[width=0.8\linewidth]{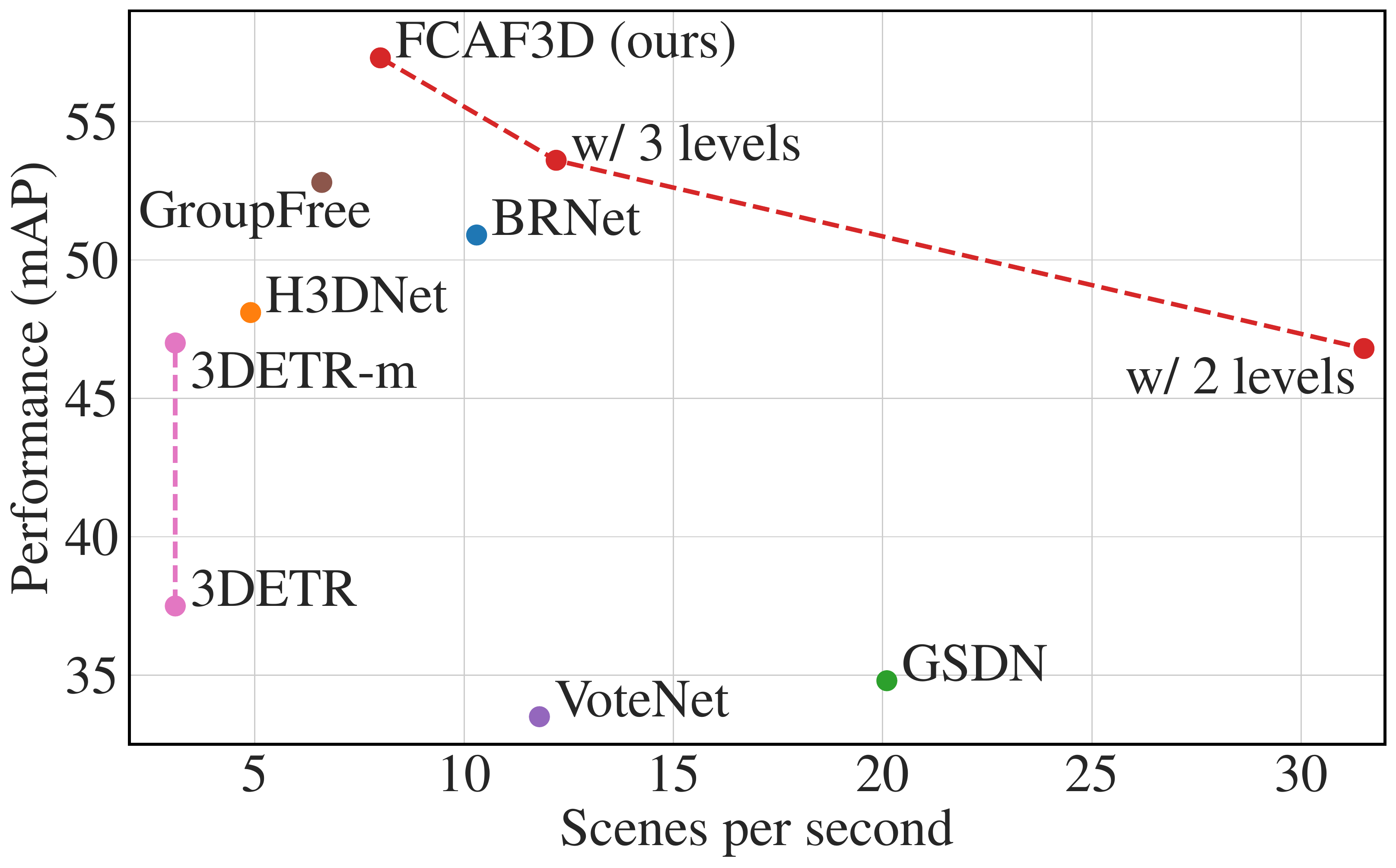}
    \caption{mAP@0.5 scores on ScanNet against scenes per second. FCAF3D modifications (marked red) have different number of backbone feature levels. For each existing method, there is a FCAF3D modification surpassing this method in both detection accuracy and inference speed.}
    \label{fig:fps}
\end{figure}

While 2D methods (\cite{tian2019fcos}, \cite{zhang2020atss}) work with dense fixed-size arrays, 3D methods are challenged by irregular unstructured 3D data of arbitrary volume. Consequently, the 2D data processing techniques are not directly applicable for 3D object detection, so 3D object detection methods (\cite{gwak2020gsdn}, \cite{qi2019votenet}, \cite{misra20213detr}) employ inventive approaches to 3D data processing. 

Convolutional 3D object detection methods have scalability issues: large-scale scenes either require an impractical amount of computational resources or take too much time to process. Other methods opt for voxel data representation and employ sparse convolutions; however, these methods solve scalability problems at the cost of detection accuracy. In other words, there is no 3D object detection method that provides precise estimates \emph{and} scales well.

Besides being scalable and accurate, an ideal 3D object detection method should handle objects of arbitrary shapes and sizes without additional hacks and hand-tuned hyperparameters. We argue that prior assumptions on 3D object bounding boxes (e.g. aspect ratios or absolute sizes) restrict generalization and increase the number of hyperparameters and trainable parameters. 

On the contrary, we do not want to rely on prior assumptions. We propose an anchor-free method that does not impose priors on objects and addresses 3D object detection with a purely data-driven approach. Moreover, we introduce a novel oriented bounding box (OBB) parametrization inspired by a Mobius strip that reduces the number of hyperparameters. To prove the effectiveness of our parametrization, we conduct experiments on SUN RGB-D with several 3D object detection methods and report improved results for all these methods.

In this paper, we present FCAF3D --- a simple, effective, and scalable method for detecting 3D objects from point clouds. We evaluate the proposed method on ScanNet \cite{dai2017scannet}, SUN RGB-D \cite{song2015sunrgbd}, and S3DIS \cite{armeni2016s3dis}, demonstrating the solid superiority over the previous state-of-the-art on all benchmarks. On SUN RGB-D and ScanNet, our method surpasses other methods by at least 3.5\% mAP@0.5. On S3DIS, FCAF3D outperforms the competitors by a huge margin.

Overall, our contribution is three-fold:
\begin{enumerate}
    \item To our knowledge, we propose a first-in-class fully convolutional anchor-free 3D object detection method (FCAF3D) for indoor scenes.
    \item We present a novel OBB parametrization and prove it to boost the accuracy of several existing 3D object detection methods on SUN RGB-D.
    \item Our method significantly outperforms the previous state-of-the-art on challenging large-scale indoor ScanNet, SUN RGB-D, and S3DIS datasets in terms of mAP while being faster on inference.
\end{enumerate}

\section{Related Work}

Recent 3D object detection methods are designed to be either indoor or outdoor. Indoor and outdoor methods have been developing almost independently, applying domain-specific data processing techniques. Many modern outdoor methods \cite{yan2018second}, \cite{lang2019pointpillars}, \cite{zhou2018voxelnet} project 3D points onto a bird-eye-view plane, thus reducing the task of 3D object detection to 2D object detection. Naturally, these methods take advantage of the fast-evolving algorithms for 2D object detection. Given a bird-eye-view projection, \cite{li20173d-fully-conv} processes it in a fully convolutional manner, while \cite{yin2021center-point} exploits 2D anchor-free approach. Unfortunately, the approaches that proved to be effective for both 2D object detection and 3D outdoor object detection cannot be trivially adapted to indoor, as it would require an impracticable amount of memory and computing resources. To address performance issues, different 3D data processing strategies have been proposed. Currently, three approaches dominate the field of 3D object detection - voting-based, transformer-based, and 3D convolutional. Below we discuss each of these approaches in detail; we also provide a brief overview of anchor-free methods.

\textbf{Voting-based methods}. VoteNet \cite{qi2019votenet} was the first method that introduced points voting for 3D object detection. VoteNet processes 3D points with PointNet \cite{qi2017pointnet}, assigns a group of points to each object candidate according to their voted center, and computes object features from each point group. Among the numerous successors of VoteNet, the major progress is associated with advanced grouping and voting strategies applied to the PointNet features. BRNet \cite{cheng2021brnet} refines voting results with the representative points from the vote centers, which improves capturing the fine local structural features. MLCVNet \cite{xie2020mlcvnet} introduces three context modules into the voting and classifying stages of VoteNet to encode contextual information at different levels. H3DNet \cite{zhang2020h3dnet} improves the point group generation procedure by predicting a hybrid set of geometric primitives. VENet \cite{xie2021venet} incorporates an attention mechanism and introduces a vote weighting module trained via a novel vote attraction loss.

All VoteNet-like voting-based methods are limited by design. First, they show poor scalability: as their performance depends on the amount of input data, they tend to slow down if given larger scenes. Moreover, many voting-based methods implement voting and grouping strategies as custom layers, making it difficult to reproduce or debug these methods or port them to mobile devices. 

\textbf{Transformer-based methods.} The recently emerged transformer-based methods use end-to-end learning and forward pass on inference instead of heuristics and optimization, which makes them less domain-specific. GroupFree \cite{liu2021group-free} replaces VoteNet head with a transformer module, updating object query locations iteratively and ensembling intermediate detection results. 3DETR \cite{misra20213detr} was the first method of 3D object detection implemented as an end-to-end trainable transformer. However, more advanced transformer-based methods still experience scalability issues similar to early voting-based methods. Differently, our method is fully-convolutional, thus being faster and significantly easier to implement than both voting-based and transformer-based methods.

\textbf{3D convolutional methods}. Voxel representation allows handling cubically growing sparse 3D data efficiently. Voxel-based 3D object detection methods~(\cite{hou20193dsis}, \cite{maturana2015voxnet}, \cite{shen2020frustum}) convert points into voxels and process them with 3D convolutional networks. However, dense volumetric features still consume much memory, and 3D convolutions are computationally expensive. Overall, processing large scenes requires a lot of resources and cannot be done within a single pass.

GSDN \cite{gwak2020gsdn} tackles performance issues with sparse 3D convolutions. It has encoder-decoder architecture, with both encoder and decoder parts built from sparse 3D convolutional blocks. Compared to the standard convolutional voting-based and transformer-based approaches, GSDN is significantly more memory-efficient and scales to large scenes without sacrificing point density. The major weakness of GSDN is its accuracy: this method is comparable to VoteNet in terms of quality, being significantly inferior to the current state-of-the-art \cite{liu2021group-free}.

\begin{figure*}[ht!]
    \centering
        \includegraphics[width=1.\linewidth]{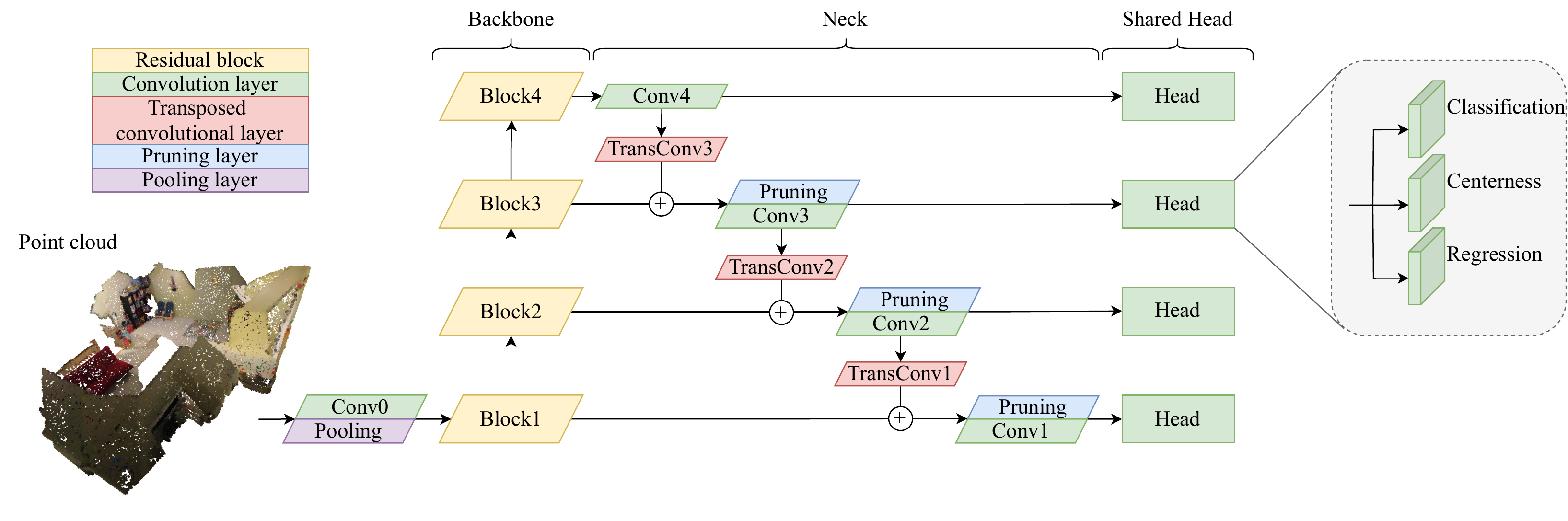}
    \caption{The general scheme of the proposed FCAF3D. All convolutions and transposed convolutions are three-dimensional and sparse. This design allows processing the input point cloud in a single forward pass.}
    \label{fig:fcaf3d}
\end{figure*}

GSDN uses 15 aspect ratios for 3D object bounding boxes as anchors. If GSDN is trained in an anchor-free setting with a single aspect ratio, the accuracy decreases by 12\%. Unlike GSDN, our method is anchor-free while taking advantage of sparse 3D convolutions.

\textbf{RGB-based anchor-free object detection.} In 2D object detection, anchor-free methods are solid competitors for the standard anchor-based methods. FCOS \cite{tian2019fcos} addresses 2D object detection in a per-pixel prediction manner and shows a robust improvement over its anchor-based predecessor RetinaNet \cite{lin2017retinanet}. FCOS3D \cite{wang2021fcos3d} trivially adapts FCOS by adding extra targets for monocular 3D object detection. ImVoxelNet \cite{rukhovich2021imvoxelnet} solves the same problem with an FCOS-like head built from standard (non-sparse) 3D convolutional blocks. We adapt the ideas from mentioned anchor-free methods to process sparse irregular data.

\section{Proposed Method}

Following the standard 3D detection problem statement, FCAF3D accepts $N_{\text{pts}}$ RGB-colored points and outputs a set of 3D object bounding boxes. The FCAF3D architecture consists of a backbone, a neck, and a head (depicted in Fig. \ref{fig:fcaf3d}).

While designing FCAF3D, we aim for scalability, so we opt for a GSDN-like sparse convolutional network. For better generalization, we reduce the number of hyperparameters in this network that need to be manually tuned; specifically, we simplify sparsity pruning in the neck. Furthermore, we introduce an anchor-free head with a simple multi-level location assignment. Finally, we discuss the limitations of existing 3D bounding box parametrizations and propose a novel parametrization that improves both accuracy and generalization ability.

\subsection{Sparse Neural Network}

\textbf{Backbone.} The backbone in FCAF3D is a sparse modification of ResNet \cite{he2016resnet} where all 2D convolutions are replaced with sparse 3D convolutions. The family of sparse high-dimensional versions of ResNet was first introduced in \cite{choy2019minkowski}; for brevity, we refer to them as to HDResNet. 

\textbf{Neck.} Our neck is a simplified GSDN decoder. Features on each level are processed with one sparse transposed 3D convolution and one sparse 3D convolution. Each transposed sparse 3D convolution with a kernel size of 2 might increase the number of non-zero values by $2^3$ times. To prevent rapid memory growth, GSDN uses the \textit{pruning} layer that filters input with a probability mask.

In GSDN, feature level-wise probabilities are calculated with an additional convolutional scoring layer. This layer is trained with a special loss encouraging consistency between the predicted sparsity and anchors. Specifically, voxel sparsity is set to be positive if any of the subsequent anchors associated with the current voxel is positive. However, using this loss may be suboptimal, as distant voxels of an object might get assigned with a low probability. 

For simplicity, we remove the scoring layer with the corresponding loss and use probabilities from the classification layer in the head instead. We do not tune the probability threshold but keep at most $N_{\text{vox}}$ voxels to control the sparsity level, where $N_{\text{vox}}$ equals the number of input points $N_{\text{pts}}$. We claim this to be a simple yet elegant way to prevent sparsity growth since reusing the same hyperparameter makes the process more transparent and consistent.


\textbf{Head.} The anchor-free FCAF3D head consists of three parallel sparse convolutional layers with weights shared across feature levels. For each location $(\hat{x}, \hat{y}, \hat{z})$, these layers output classification probabilities $\hat{\boldsymbol{p}}$, bounding box regression parameters $\boldsymbol{\delta}$, and centerness $\hat{c}$, respectively. This design is similar to the simple and light-weight head of FCOS \cite{tian2019fcos} but adapted to 3D data.

\textbf{Multi-level location assignment.} During training, FCAF3D outputs locations $\{(\hat{x},\hat{y},\hat{z})\}$ for different feature levels, which should be assigned to ground truth boxes $\{\boldsymbol{b}\}$. For each location, FCOS \cite{tian2019fcos} and ImVoxelNet \cite{rukhovich2021imvoxelnet} consider ground truth bounding boxes covering this location, whose faces are all within distance threshold, select the bounding box with the least volume, and assign it to this location. Such a strategy is suboptimal, and its alterations are widely explored in 2D object detection \cite{zhang2020atss}, \cite{ge2021ota}. ImVoxelNet \cite{rukhovich2021imvoxelnet} uses a modified strategy that requires hand-tuning the face distance threshold for each feature level.

We propose a simplified strategy for sparse data that does not require tuning dataset-specific hyperparameters. For each bounding box, we select the last feature level at which this bounding box covers at least $N_{\text{loc}}$ locations. If there is no such a feature level, we opt for the first one. We also filter locations via \textit{center sampling}~\cite{tian2019fcos}, considering only the points near the bounding box center as positive matches. More details are presented in Sec. \ref{subsec:ablation}.

Through assignment, some locations $\{(\hat{x}, \hat{y}, \hat{z})\}$ are matched with ground truth bounding boxes $\boldsymbol{b}_{\hat{x},\hat{y},\hat{z}}$. Accordingly, these locations get associated with ground truth labels $p_{\hat{x},\hat{y},\hat{z}}$ and 3D centerness values $c_{\hat{x},\hat{y},\hat{z}}$. During inference, the scores $\hat{\boldsymbol{p}}$ are multiplied by 3D centerness $\hat{c}$ just before NMS as proposed in  \cite{rukhovich2021imvoxelnet}.

\textbf{Loss function.} The overall loss function is formulated as follows:
\begin{equation}\begin{split}
    L=\frac{1}{N_\text{pos}}\sum_{\hat{x},\hat{y},\hat{z}}(L_\text{cls}(\hat{\boldsymbol{p}}, p)
    + \mathbb{1}_{\{p_{\hat{x},\hat{y},\hat{z}} \neq 0\}} L_\text{reg}(\hat{\boldsymbol{b}}, \boldsymbol{b})
    + \mathbb{1}_{\{p_{\hat{x},\hat{y},\hat{z}} \neq 0\}} L_\text{cntr}(\hat{c}, c)).
\end{split}\end{equation}
Here, the number of matched locations $N_{\text{pos}}$ is $\sum_{\hat{x},\hat{y},\hat{z}} \mathbb{1}_{\{p_{\hat{x},\hat{y},\hat{z}} \neq 0\}}$. Classification loss $L_\text{cls}$ is a focal loss, regression loss $L_\text{reg}$ is IoU, and centerness loss $L_\text{cntr}$ is binary cross-entropy. For each loss, predicted values are denoted with a hat.

\subsection{Bounding Box Parametrization}

The 3D object bounding boxes can be axis-aligned (AABB) or oriented (OBB). An AABB can be described as $\boldsymbol{b}^{\text{AABB}}=(x, y, z, w, l, h)$, while the definition of an OBB includes a \textit{heading angle} $\theta$: $\boldsymbol{b}^{\text{OBB}}=(x, y, z, w, l, h, \theta)$. In both formulas, $x, y, z$ denote the coordinates of the center of a bounding box, while $w, l, h$ are its width, length, and height, respectively.

\textbf{AABB parametrization}. For AABBs, we follow the parametrization proposed in \cite{rukhovich2021imvoxelnet}. Specifically, for a ground truth AABB $(x, y, z, w, l, h)$ and a location $(\hat{x}, \hat{y}, \hat{z})$, $\boldsymbol{\delta}$ can be formulated as a 6-tuple:
\begin{equation}\label{eq:delta1-6}\begin{split}
    \delta_1=x + \frac{w}{2} - \hat{x},\ \delta_2=\hat{x} - x + \frac{w}{2},\ \delta_3=y + \frac{l}{2} - \hat{y}, \\
    \delta_4=\hat{y} - y + \frac{l}{2},\ \delta_5=z + \frac{h}{2} - \hat{z},\ \delta_6=\hat{z} - z + \frac{h}{2}.
\end{split}\end{equation}
The predicted AABB $\boldsymbol{\hat{b}}$ can be trivially obtained from $\boldsymbol{\delta}$.

\textbf{Heading angle estimation.} All state-of-the-art 3D object detection methods from point clouds address the heading angle estimation task as classification followed by regression. The heading angle is classified into bins; then, the precise heading angle is regressed within a bin. For indoor scenes, the range from 0 to $2 \pi$ is typically divided into 12 equal bins~\cite{qi2019votenet}, \cite{qi2020imvotenet}, \cite{zhang2020h3dnet}, \cite{misra20213detr}. For outdoor scenes, there are usually only two bins~\cite{yan2018second}, \cite{lang2019pointpillars}, as the objects on the road can be either parallel or perpendicular to the road.

When a heading angle bin is chosen, the heading angle value is estimated through regression. VoteNet and other voting-based methods estimate the value of $\theta$ directly. Outdoor methods explore more elaborate approaches, e.g. predicting the values of trigonometric functions. For instance, SMOKE \cite{liu2020smoke} estimates $\sin \theta$ and $\cos \theta$ and uses the predicted values to recover the heading angle. 

\begin{wrapfigure}{r}{0.6\textwidth}
\setlength{\tabcolsep}{2pt}
\begin{tabular}{ccc}
    \includegraphics[height=60pt]{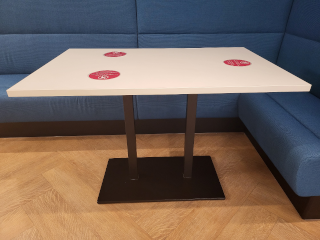} &
    \includegraphics[height=60pt]{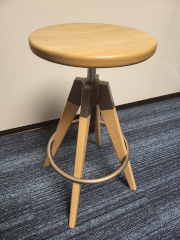} &
    \includegraphics[height=60pt]{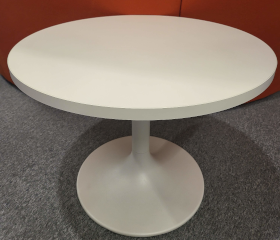}
\end{tabular}
\caption{Examples of objects with an ambiguous heading angle.}
\label{fig:direction_examples}
\end{wrapfigure}


Fig. \ref{fig:direction_examples} depicts indoor objects where the heading angle is unambiguous. Accordingly, ground truth angle annotations can be chosen randomly for these objects, making heading angle bin classification meaningless. To avoid penalizing the correct predictions that do not coincide with annotations, we use rotated IoU loss, as its value is the same for all possible choices of heading angle. Thus, we propose OBB parametrization that considers the rotation ambiguity.  

\textbf{Proposed Mobius OBB parametrization}. Considering the OBB with parameters $(x, y, z, w, l, h, \theta)$, let us denote $q=\frac{w}{l}$. If $x,\ y,\ z,\ w + l,\ h$ are fixed, it turns out that the OBBs with 
\begin{equation}\label{eq:q_theta}
    \left( q, \theta \right),\ \left( \frac{1}{q},\ \theta+\frac{\pi}{2} \right),\ \left( q, \theta+\pi \right),\ \left( \frac{1}{q}, \theta+\frac{3\pi}{2} \right)
\end{equation}
define the same bounding box.
We notice that the set of $(q, \theta)$, where $\theta \in ( 0, 2\pi ], \ q \in (0, +\inf)$ is topologically equivalent to a Mobius strip \cite{munkres2000topology} up to this equivalence relation. Hence, we can reformulate the task of estimating $(q, \theta)$ as a task of predicting a point on a Mobius strip. A natural way to embed a Mobius strip being a two-dimensional manifold to Euclidean space is the following: 
\begin{equation} \label{eq:embedding}
    (q, \theta)\ \mapsto\ ( \ln(q) \sin(2\theta), \ln(q) \cos(2\theta), \sin(4\theta), \cos(4\theta)).
\end{equation}
It is easy to verify that 4 points from Eq. \ref{eq:q_theta} are mapped into a single point in Euclidean space (see Supplementary for details). 
However, the experiments reveal that predicting only $\ln(q) \sin(2\theta)$ and $\ln(q) \cos(2\theta)$ provides better results than predicting all four values. Thereby, we opt for a \textit{pseudo} embedding of a Mobius strip to $\mathbb{R}^2$. We call it \textit{pseudo} since it maps the entire center circle of a Mobius strip defined by $\ln(q)=0$ to $(0, 0)$. Accordingly, we cannot distinguish points with $\ln q=0$. However, $\ln(q)=0$ implies strict equality of $w$ and $l$, which is rare in real-world scenarios. Moreover, the choice of an angle has a minor effect on the IoU if $w = l$; thereby, we ignore this rare case for the sake of detection accuracy and simplicity of the method. Overall, we obtain a novel OBB parametrization: 
\begin{equation}\label{eq:delta7-8}
    \delta_7=\ln\frac{w}{l}\sin(2\theta),\ \delta_8=\ln\frac{w}{l}\cos(2\theta).
\end{equation}

In the standard parametrization \ref{eq:delta1-6}, $\hat{\boldsymbol{b}}$ is trivially derived from $\boldsymbol{\delta}$. In the proposed parametrization, $w, l, \theta$ are non-trivial and can be obtained as follows:
\begin{equation}
    w=\frac{sq}{1+q},\ l=\frac{s}{1+q},\ \theta=\frac{1}{2}\arctan{\frac{\delta_7}{\delta_8}},
\end{equation}
where ratio $q=e^{\sqrt{\delta_7^2+\delta_8^2}}$ and size $s=\delta_1+\delta_2+\delta_3+\delta_4$.

\section{Experiments}

\subsection{Datasets}

We evaluate our method on three 3D object detection benchmarks: ScanNet V2 \cite{dai2017scannet}, SUN RGB-D \cite{song2015sunrgbd}, and S3DIS \cite{armeni2016s3dis}. For all datasets, we use mean average precision (mAP) under IoU thresholds of 0.25 and 0.5 as a metric.

\textbf{ScanNet.} The ScanNet dataset contains 1513 reconstructed 3D indoor scans with per-point instance and semantic labels of 18 object categories. Given this annotation, we calculate AABBs via the standard approach \cite{qi2019votenet}. The training subset is comprised of 1201 scans, while 312 scans are left for validation.

\textbf{SUN RGB-D.} SUN RGB-D is a monocular 3D scene understanding dataset containing more than 10,000 indoor RGB-D images. The annotation consists of per-point semantic labels and OBBs of 37 object categories. As proposed in \cite{qi2019votenet}, we run experiments with objects of the 10 most common categories. The training and validation splits contain 5285 and 5050 point clouds, respectively.

\textbf{S3DIS.} Stanford Large-Scale 3D Indoor Spaces dataset contains 3D scans of 272 rooms from 6 buildings, with 3D instance and semantic annotation. Following \cite{gwak2020gsdn}, we evaluate our method on furniture categories. AABBs are derived from 3D semantics. We use the official split, where 68 rooms from \textit{Area 5} are intended for validation, while the remaining 204 rooms comprise the training subset.

\subsection{Implementation Details}

\textbf{Hyperparameters.} For all datasets, we use the same hyperparameters except for the following. First, the size of output classification layer equals the number of object categories, which is 18, 10, and 5 for ScanNet, SUN RGB-D, and S3DIS. Second, SUN RGB-D contains OBBs, so we predict additional targets $\delta_7$ and $\delta_8$ for this dataset; note that the loss function is not affected. Last, ScanNet, SUN RGB-D, and S3DIS contain different numbers of scenes, so we repeat each scene 10, 3, and 13 times per epoch, respectively. 

Similar to GSDN \cite{gwak2020gsdn}, we use the sparse 3D modification of ResNet34 named HDResNet34 as a backbone. The neck and the head use the outputs of the backbone at all feature levels. In initial point cloud voxelization, we set the voxel size to 0.01m and the number of points $N_\text{pts}$ to 100,000. Respectively, $N_\text{vox}$ equals to 100,000. Both ATSS \cite{zhang2020atss} and FCOS \cite{tian2019fcos} set $N_{\text{loc}}$ to $3^2$ for 2D object detection. Accordingly, we select a feature level so bounding box covers at least $N_{\text{loc}} = 3^3$ locations. We select 18 locations by center sampling. The NMS IoU threshold is 0.5.

\textbf{Training.} We implement FCAF3D using the MMdetection3D \cite{2020mmdetection3d} framework. The training procedure follows the default MMdetection \cite{chen2019mmdetection} scheme: training takes 12 epochs and the learning rate decreases on the 8th and the 11th epochs. We employ the Adam optimizer with an initial learning rate of 0.001 and weight decay of 0.0001. All models are trained on two NVidia V100 with a batch size of 8. Evaluation and performance tests are run on a single NVidia GTX1080Ti.

\begin{table*}[h!]
\centering \setlength{\tabcolsep}{2.5pt}
    \resizebox{1\linewidth}{!}{
    \begin{tabular}{l|c|cccccc}
    \hline
    \multirow[l]{2}{*}{Method} & \multirow[c]{2}{*}{Presented at} & \multicolumn{2}{c}{ScanNet} & \multicolumn{2}{c}{SUN RGB-D} & \multicolumn{2}{c}{S3DIS} \\
    & & mAP@0.25 & mAP@0.5 & mAP@0.25 & mAP@0.5 & mAP@0.25 & mAP@0.5 \\ \hline
    VoteNet\cite{qi2019votenet} & ICCV'19 & 58.6 & 33.5 & 57.7 & - & - & - \\
    3D-MPA\cite{engelmann20203d-mpa} & CVPR'20 & 64.2 & 49.2 & - & - & - & - \\
    HGNet\cite{chen2020hgnet} & CVPR'20 & 61.3 & 34.4 & 61.6 & - & - & - \\
    MLCVNet\cite{xie2020mlcvnet} & CVPR'20 & 64.5 & 41.4 & 59.8 & - & - & - \\
    GSDN\cite{gwak2020gsdn} & ECCV'20 & 62.8 & 34.8 & - & - & 47.8 & 25.1 \\
    H3DNet\cite{zhang2020h3dnet} & ECCV'20 & 67.2 & 48.1 & 60.1 & 39.0 & - & - \\
    BRNet\cite{cheng2021brnet} & CVPR'21 & 66.1 & 50.9 & 61.1 & 43.7 & - & - \\
    3DETR\cite{misra20213detr} & ICCV'21 & 65.0 & 47.0 & 59.1 & 32.7 & - & - \\
    VENet\cite{xie2021venet} & ICCV'21 & 67.7 & - & 62.5 & 39.2 & - & - \\
    GroupFree \cite{liu2021group-free} & ICCV'21 & 69.1 (68.6) & 52.8 (51.8) & 63.0 (62.6) & 45.2 (44.4) & - & - \\
    FCAF3D & - & \textbf{71.5} (70.7) & \textbf{57.3} (56.0) & \textbf{64.2} (63.8) & \textbf{48.9} (48.2) & \textbf{66.7} (64.9) & \textbf{45.9} (43.8) \\ \hline
    \end{tabular}
    }
    \caption{Results of FCAF3D and existing indoor 3D object detection methods that accept point clouds. The best metric values are marked bold. FCAF3D outperforms previous state-of-the-art methods: GroupFree (on ScanNet and SUN RGB-D) and GSDN (on S3DIS). The reported metric value is the best one across 25 trials; the average value is given in brackets.}
    \label{tab:results}
\end{table*}

\textbf{Evaluation}. We follow the evaluation protocol introduced in \cite{liu2021group-free}. Both training and evaluation are randomized, as the input $N_\text{pts}$ are randomly sampled from the point cloud. To obtain statistically significant results, we run training 5 times and test each trained model 5 times independently. We report both the best and average metrics across $5 \times 5$ trials: this allows comparing FCAF3D to the 3D object detection methods that report either a single best or an average value.

\section{Results}

\subsection{Comparison with State-of-the-art Methods}

\begin{figure}[h!]
\centering
\setlength{\tabcolsep}{2pt}
\begin{tabular}{cc}
    \includegraphics[width=0.4\linewidth]{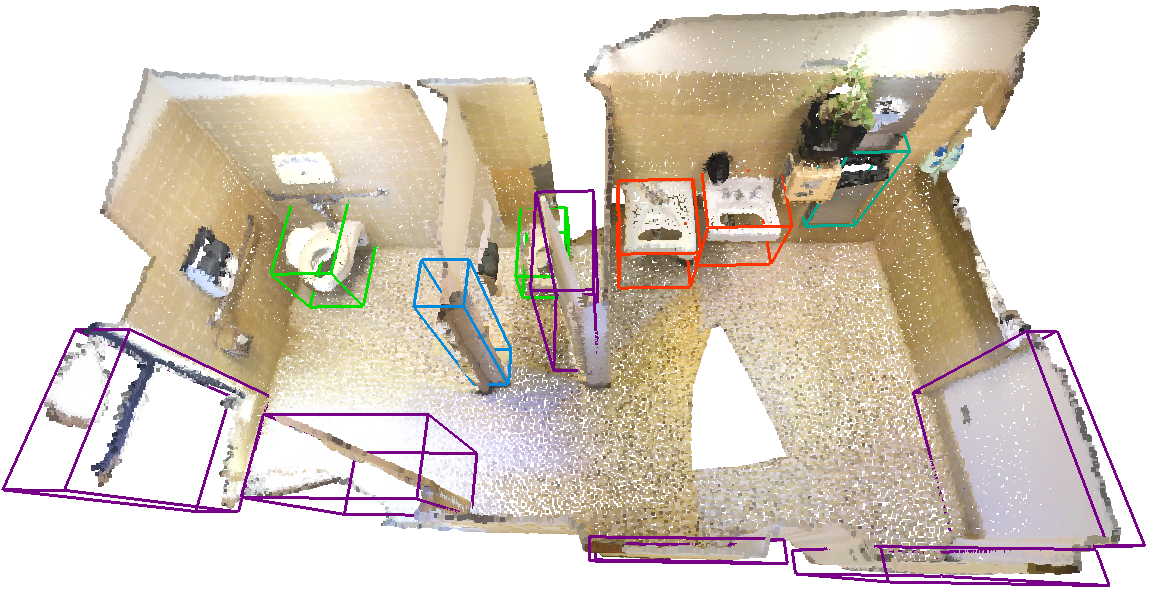} &
    \includegraphics[width=0.4\linewidth]{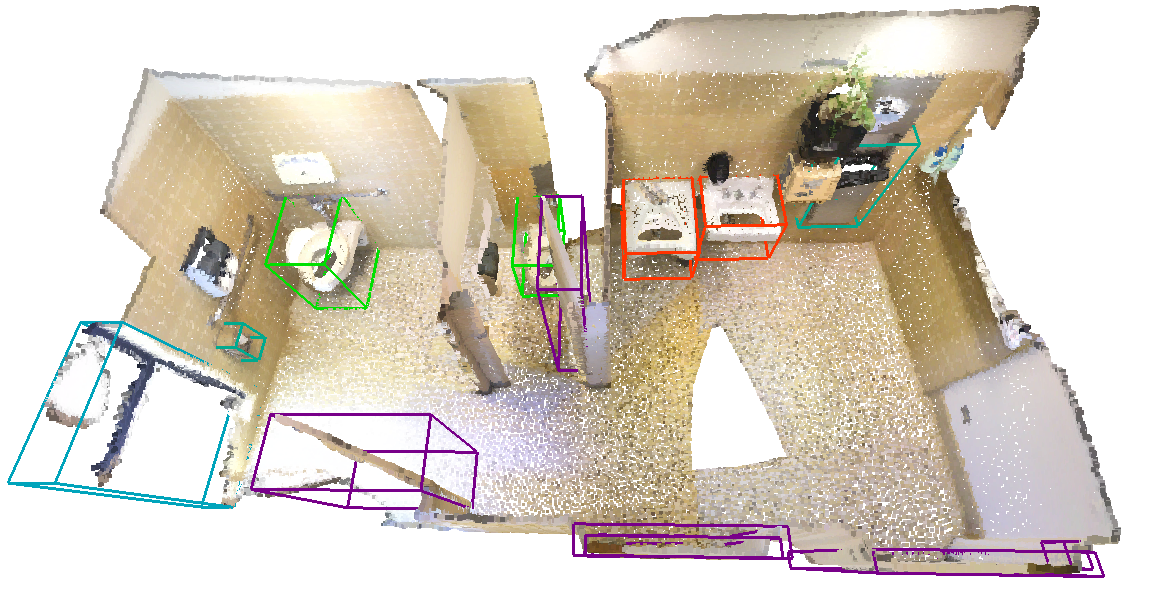} \\
    \includegraphics[width=0.4\linewidth]{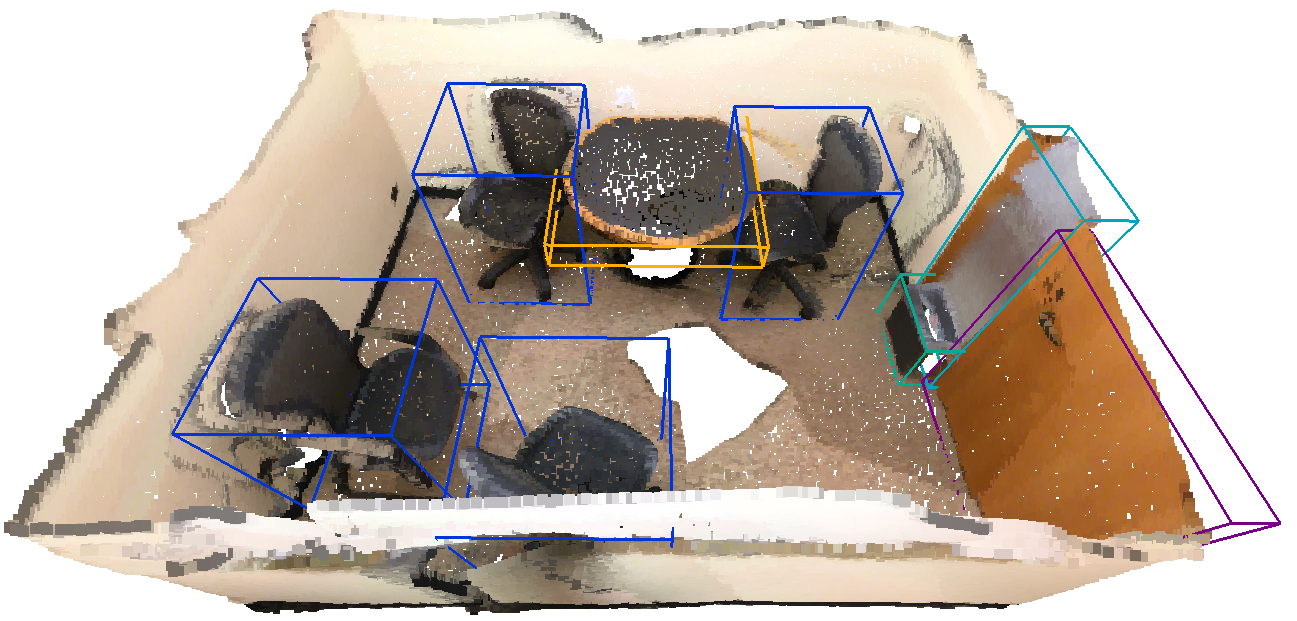} &
    \includegraphics[width=0.4\linewidth]{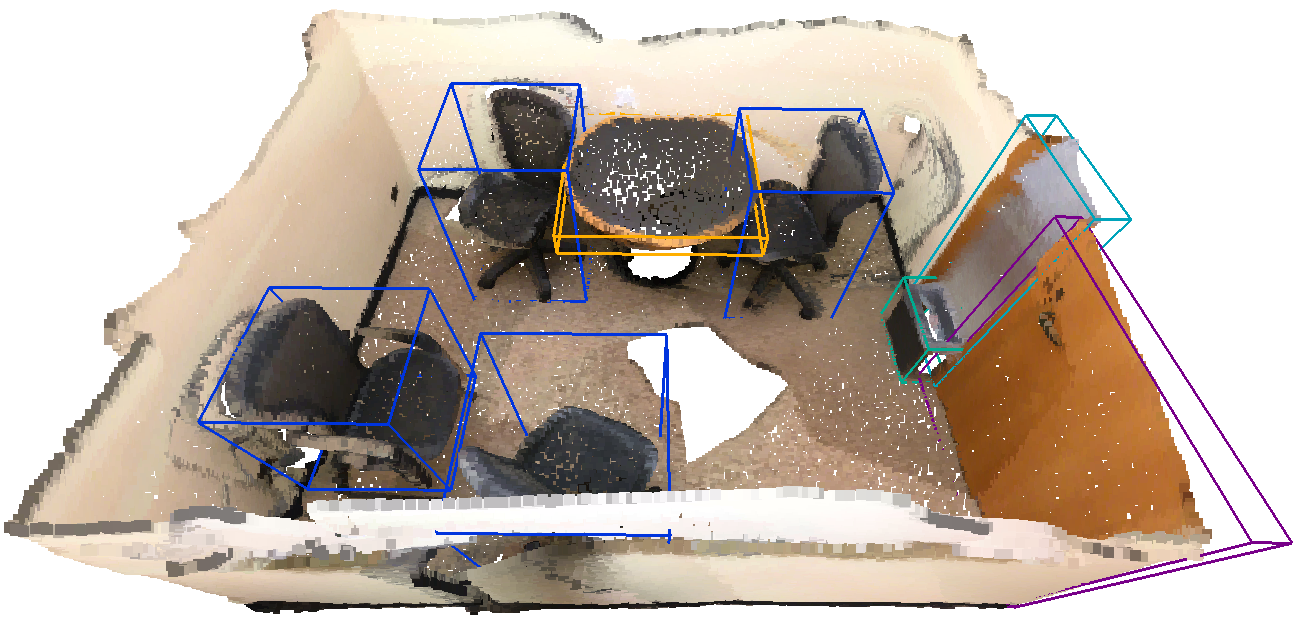}
\end{tabular}
\caption{The point cloud from ScanNet with AABBs. The color of a bounding box denotes the object category. Left: estimated with FCAF3D, right: ground truth.}
\label{fig:scannet_examples}
\end{figure}

\begin{figure}[h!]
\centering
\setlength{\tabcolsep}{2pt}
\begin{tabular}{cc}
    \includegraphics[width=0.4\linewidth]{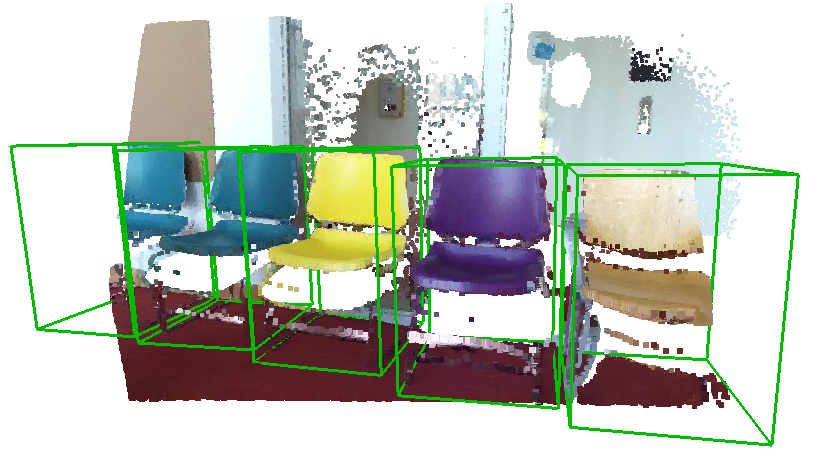} &
    \includegraphics[width=0.4\linewidth]{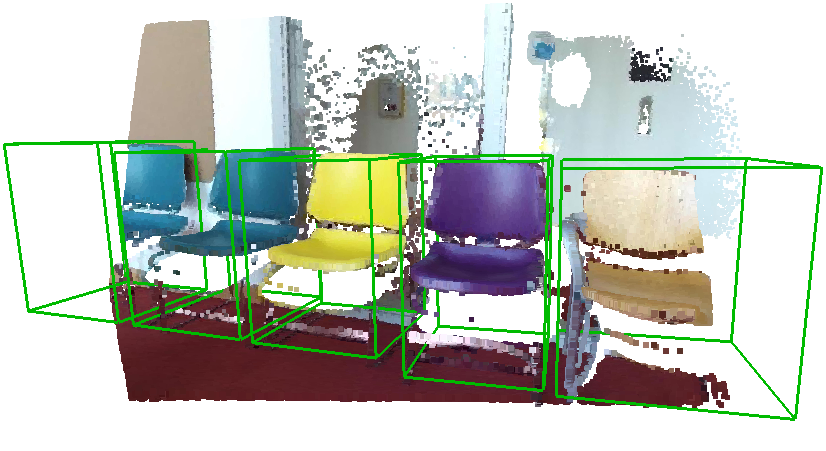} \\
    \includegraphics[width=0.4\linewidth]{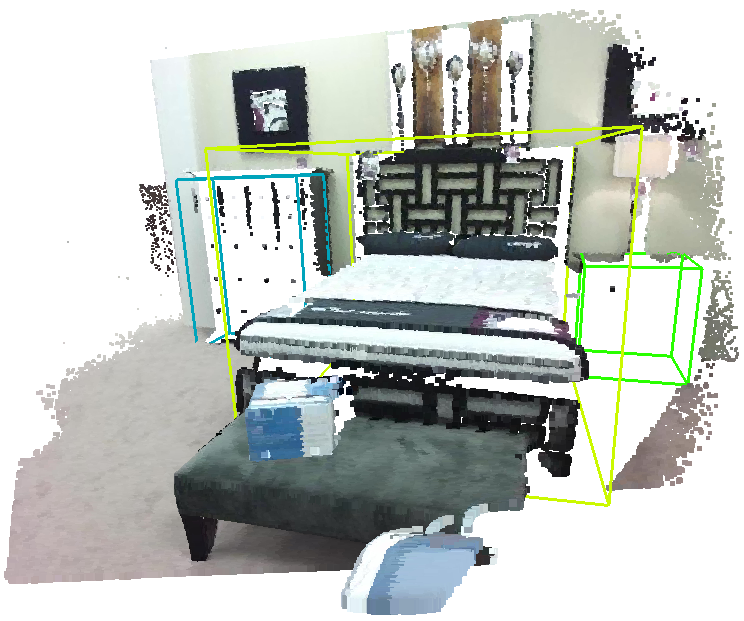} &
    \includegraphics[width=0.4\linewidth]{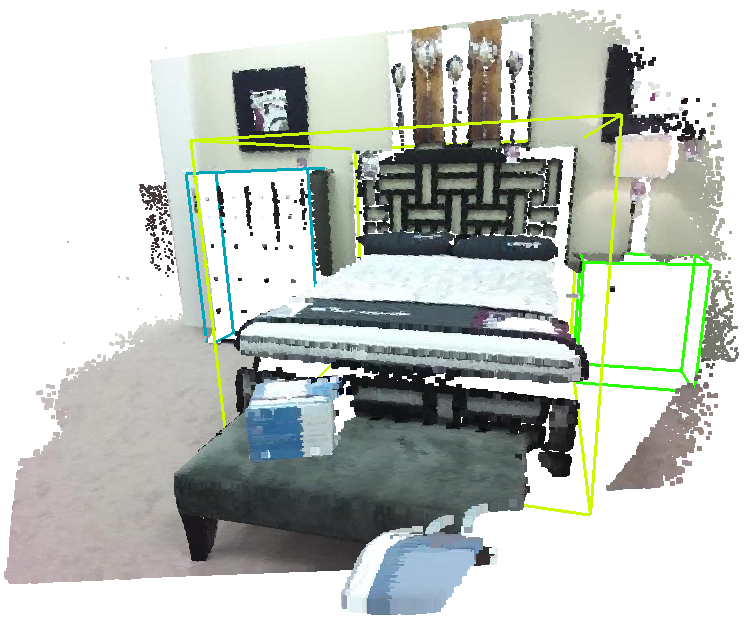}
\end{tabular}
\caption{The point cloud from SUN RGB-D with OBBs. The color of a bounding box denotes the object category. Left: estimated with FCAF3D, right: ground truth.}
\label{fig:sunrgbd_examples}
\end{figure}

\begin{figure}[h!]
\centering
\setlength{\tabcolsep}{2pt}
\begin{tabular}{cc}
    \includegraphics[width=0.4\linewidth]{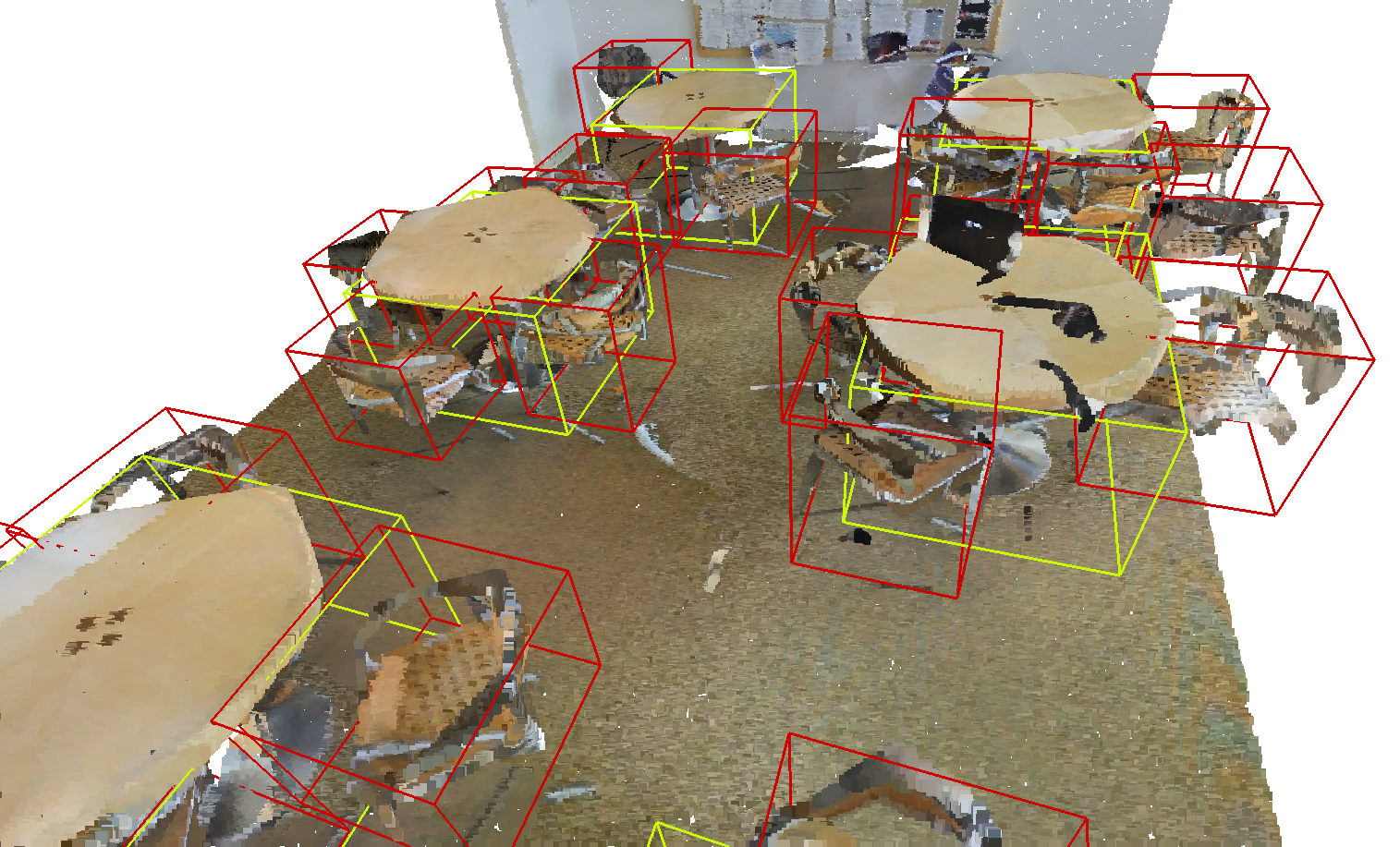} &
    \includegraphics[width=0.4\linewidth]{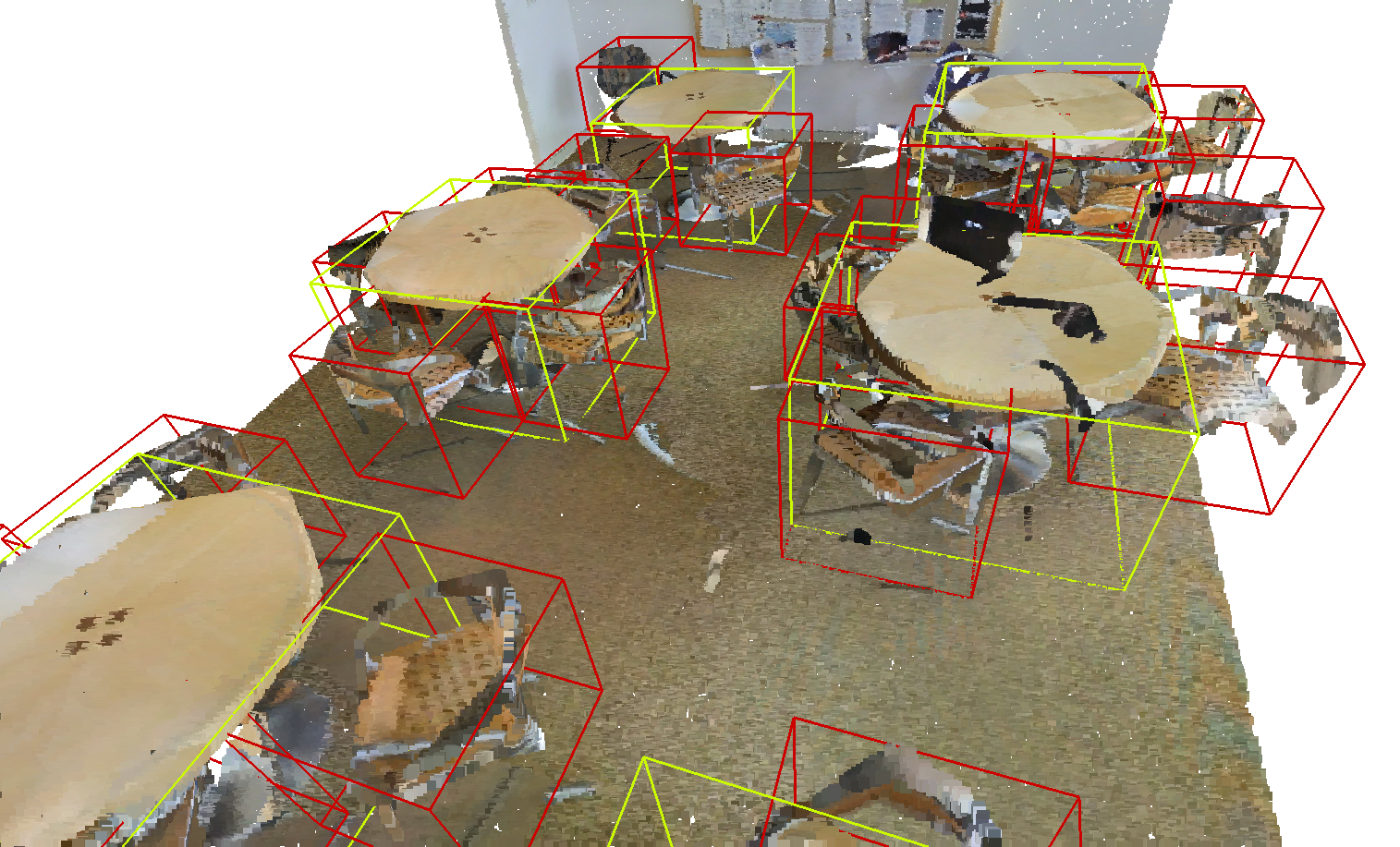} \\
    \includegraphics[width=0.4\linewidth]{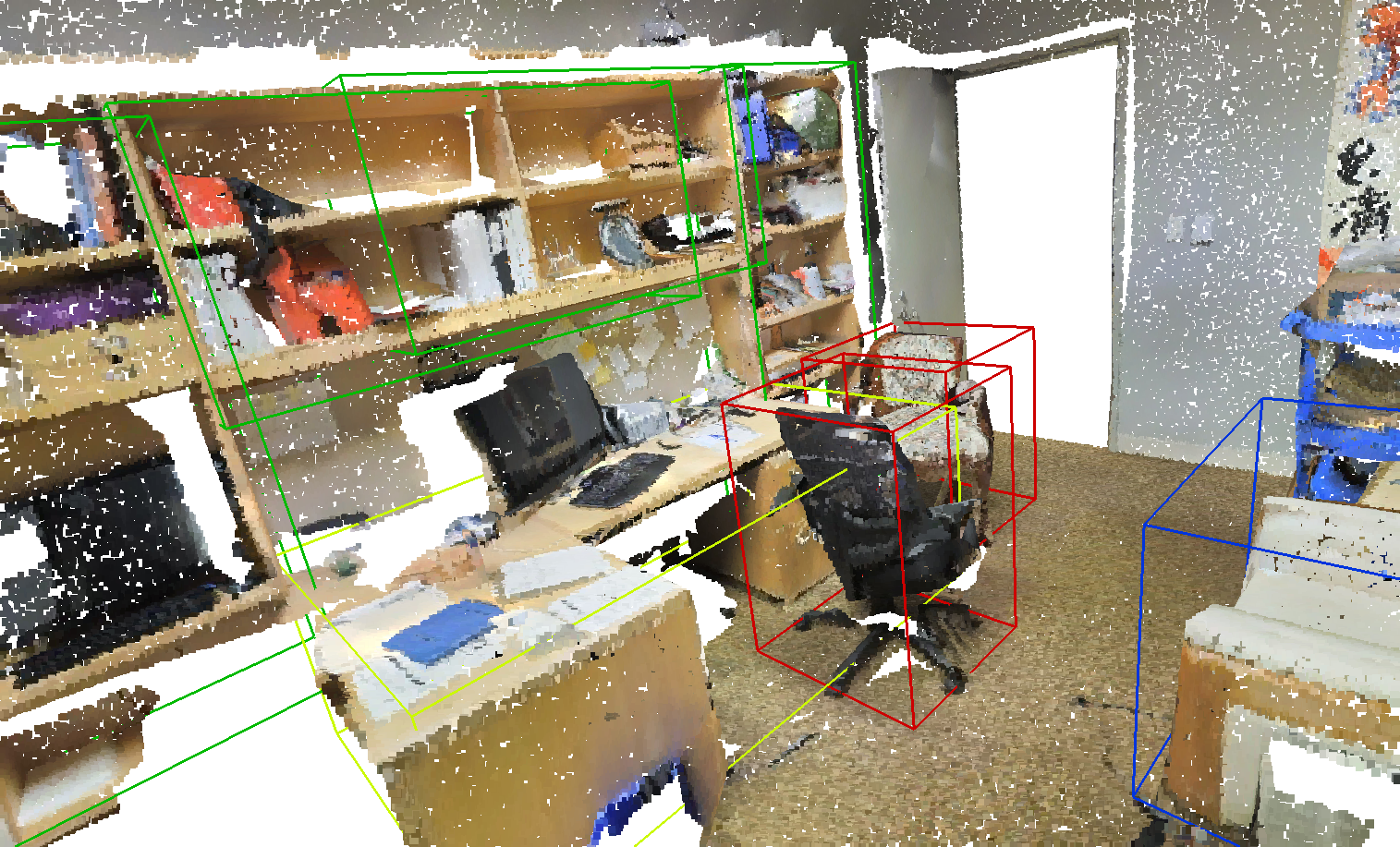} &
    \includegraphics[width=0.4\linewidth]{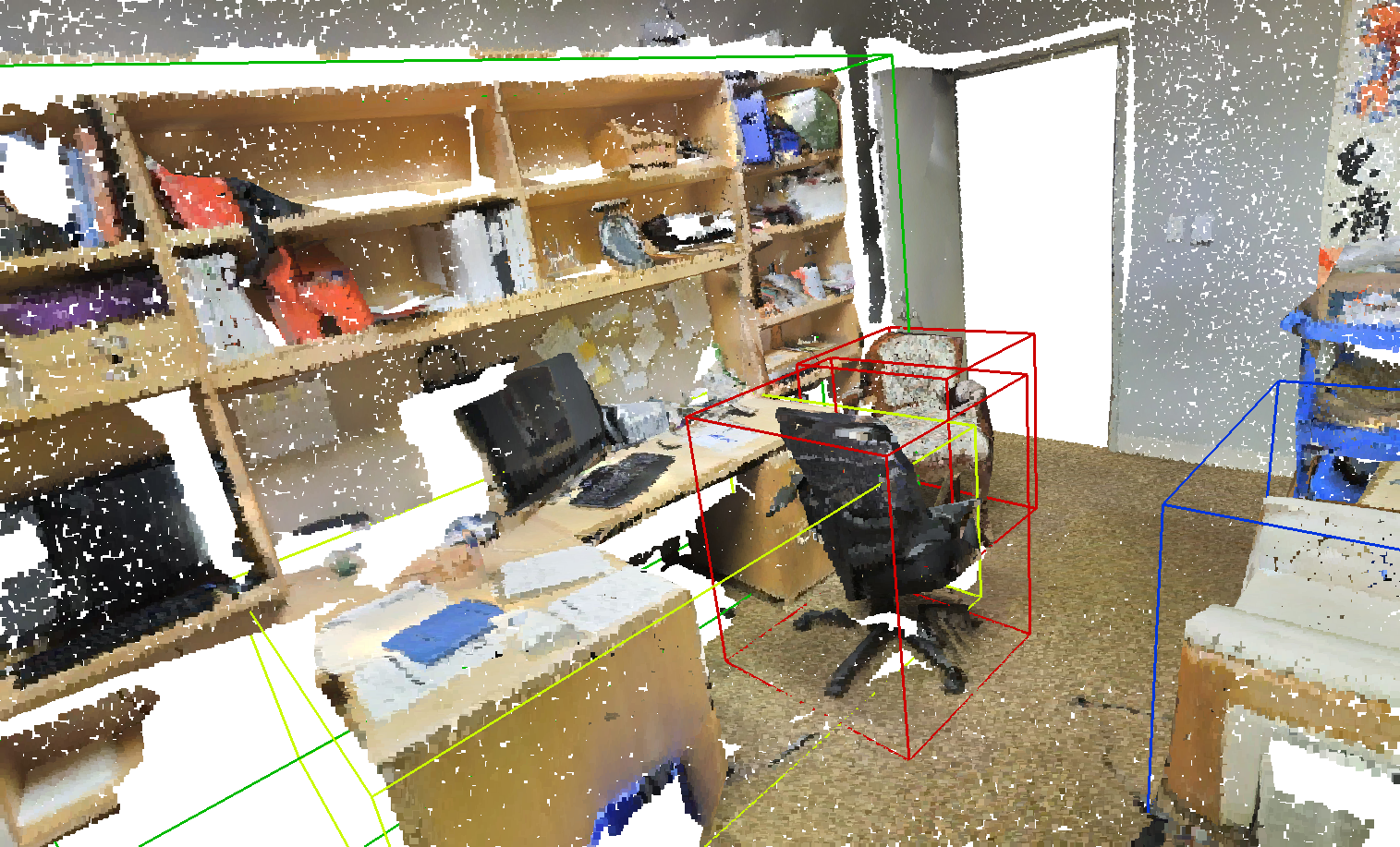}
\end{tabular}
\caption{The point cloud from S3DIS with AABBs. The color of a bounding box denotes the object category. Left: estimated with FCAF3D, right: ground truth.}
\label{fig:s3dis_examples}
\end{figure}

We compare FCAF3D with previous state-of-the-arts on three indoor benchmarks in Tab. \ref{tab:results}. As one might observe, FCAF3D achieves the best results on all benchmarks. The performance gap is especially tangible in terms of mAP@0.5: our method surpasses previous state-of-the-art by 4.5\% on ScanNet and 3.7\% on SUN RGB-D. On S3DIS, FCAF3D outperforms weak state-of-the-art by a huge margin. Overall, the proposed method is consistently better than existing methods, setting a new state-of-the-art for indoor 3D object detection. The examples of ScanNet, SUN RGB-D, and S3DIS point clouds with predicted bounding boxes are depicted in Fig. \ref{fig:scannet_examples}, \ref{fig:sunrgbd_examples}, \ref{fig:s3dis_examples}.

\subsection{Object Geometry Priors}

To study geometry priors, we train and evaluate existing methods with proposed modifications. We experiment with 3D object detection methods accepting data of different modalities: point clouds, RGB images, or both, to see whether the effect is data-specific or universal. VoteNet and ImVoteNet have the same head and are trained with the same losses. Among them, there are 4 prior losses: size classification loss, size regression loss, direction classification loss, and direction regression loss. Both classification losses correspond to targets parametrized using priors (per-category mean object sizes and a set of angle bins). Similar to FCAF3D, we replace the aforementioned losses with a rotated IoU loss with Mobius parametrization \ref{eq:delta7-8}. To give a complete picture, we also try a \textit{sin-cos} parametrization used in the outdoor 3D object detection method SMOKE \cite{liu2020smoke}.

The rotated IoU loss decreases the number of trainable parameters and hyperparameters, including geometry priors and loss weights. This loss has already been used in outdoor 3D object detection \cite{zhou2019iou}. Recently, \cite{2020mmdetection3d} reported results of VoteNet trained with axis-aligned IoU loss on ScanNet.

Tab. \ref{tab:ablation_priors} shows that replacing the standard parametrization with Mobius one boosts VoteNet and ImVoteNet mAP@0.5 by approximately 4\%. 

ImVoxelNet does not use a classification+regression scheme to estimate heading angle but predicts its value directly in a single step. Since the original ImVoxelNet uses the rotated IoU loss, we do not need to remove redundant losses, only to change the parametrization. Again, the Mobius parametrization helps to obtain the best results, even though the superiority is minor.

\begin{table}[h!]
    \centering \setlength{\tabcolsep}{2.5pt}
    \begin{tabular}{l|c|cc}
    \hline
    Method & Input & mAP@0.25 & mAP@0.5 \\ \hline
    VoteNet\cite{qi2019votenet} & \multirow{6}{*}{PC} & 57.7 & - \\
    \ \ Reimpl.\cite{2020mmdetection3d} & & 59.1 & 35.8 \\
    \ \ Reimpl. w/ IoU loss & & & \\
    \ \ \ \ w/ naive param. & & 61.1 (60.3) & 38.4 (37.7) \\
    \ \ \ \ w/ \textit{sin-cos} param. & & 60.7 (59.8) & 37.1 (36.4) \\
    \ \ \ \ w/ Mobius param. & & \textbf{61.1} (60.5) & \textbf{40.4} (39.5) \\ \hline
    ImVoteNet\cite{qi2020imvotenet} & \multirowcell{6}{RGB \\ +PC} & 63.4 & - \\
    \ \ Reimpl.\cite{2020mmdetection3d} & & 64.0 & 37.8 \\
    \ \ Reimpl. w/ IoU loss & & & \\
    \ \ \ \ w/ naive param. & & 64.2 (63.9) & 39.1 (38.3) \\
    \ \ \ \ w/ \textit{sin-cos} param. & & 64.6 (64.0) & 39.9 (37.8) \\
    \ \ \ \ w/ Mobius param. & & \textbf{64.6} (64.1) & \textbf{40.8} (39.8) \\ \hline
    ImVoxelNet\cite{rukhovich2021imvoxelnet} & \multirow{4}{*}{RGB} & 40.7 & - \\
    \ \ w/ naive param. & & 41.3 (40.4) & 13.8 (13.0) \\
    \ \ w/ \textit{sin-cos} param. & & 41.3 (40.5) & 13.2 (12.8) \\
    \ \ w/ Mobius param. & & \textbf{41.5} (40.6) & \textbf{14.6} (14.0) \\ \hline
    FCAF3D & \multirow{4}{*}{PC} & & \\
    \ \ w/ naive param. & & 63.8 (63.5) & 46.8 (46.2) \\
    \ \ w/ \textit{sin-cos} param. & & 63.9 (63.6) & 48.2 (47.3) \\
    \ \ w/ Mobius param. & & \textbf{64.2} (63.8) & \textbf{48.9} (48.2) \\ \hline
    \end{tabular}
    \caption{Results of several 3D object detection methods that accept inputs of different modalities, with different OBB parametrization on SUN RGB-D. The FCAF3D metric value is the best across 25 trials; the average value is given in brackets. For other methods, we report results from the original papers and also the results obtained through our experiments with MMdetection3D-based re-implementations (marked as Reimpl). PC stands for point cloud.}
    \label{tab:ablation_priors}
\end{table}

\begin{table}[h!]
    \centering \setlength{\tabcolsep}{2.5pt}
    \resizebox{1\linewidth}{!}{
    \begin{tabular}{l|cccc|c|cc}
    \hline
    \multirow[l]{2}{*}{Method} & \multirow[l]{2}{*}{Backbone} & Voxel & \multirow[l]{2}{*}{Stride} & Feature level & Scenes & \multicolumn{2}{c}{mAP} \\
    & & size [m] & & voxel sizes [m] & per sec. & 0.25 & 0.5 \\ \hline
    GSDN\cite{gwak2020gsdn} & \multirow[c]{3}{*}{HDResNet34} & 0.05 & 1 & \multirow[c]{3}{*}{0.2,0.4,0.8,1.6} & 20.1 & 62.8 & 34.8 \\
    \ \ w/o anchors & & 0.05 & 1 & & 20.4 & 56.3 & 22.7 \\
    FCAF3D & & 0.05 & 1 & & 17.0 & \textbf{64.2} & \textbf{46.2} \\
    \hline
    FCAF3D \textit{(accurate)} & HDResNet34 & 0.01 & 2 & 0.08,0.16,0.32,0.64 & 8.0 & \underline{\textbf{70.7}} & \underline{\textbf{56.0}} \\ \hline
    FCAF3D \textit{(balanced)} & HDResNet34:3 & 0.05 & 1 & 0.2,0.4,0.8 & \textbf{22.9} & \textbf{62.9} & \textbf{43.9} \\ \hline
    FCAF3D \textit{(fast)} & HDResNet34:2 & 0.02 & 2 & 0.16,0.32 & \underline{\textbf{31.5}} & \textbf{63.1} & \textbf{46.8} \\ \hline
    \end{tabular}
    }
    \caption{Results of fully convolutional 3D object detection methods that accept point clouds on ScanNet. The FCAF3D results better than the results of the original GSDN (with anchors) are marked bold. The all-best results are underlined.}
    \label{tab:ablation_gsdn}
\end{table}

\textbf{GSDN anchors}. In this study, we provide a more comprehensive comparison against GSDN and report the results in Tab.~\ref{tab:ablation_gsdn}. A fair comparison implies that we should test our method in the most similar scenario with the same set of hyperparameters. Accordingly, we use a voxel size of 0.05m, ensuring we operate the same inputs and do not benefit from using more detailed and informative spatial information. With the same input voxel size, the voxel sizes at different feature levels of the decoder are also of the same sizes (0.2, 0.4, 0.8, 1.6). 

Moreover, we introduce a minor modification to our FCAF3D network. The first 3D convolution in the network has the stride of 2 in the original FCAF3D, but in GSDN, it equals to 1. With the same stride of 1, the same voxel size and the same voxel sizes at different feature levels, FCAF3D slightly outperforms GSDN in terms of mAP@0.25 (64.2 against 62.8), while demonstrating a notable accuracy gain in mAP@0.5 (46.2 against 34.8). The number of scenes processed in a second by both these methods is comparable: it equals to 17 and 20, respectively. This minor difference in speed between FCAF3D based on HDResNet34 and GSDN is attributed to the different sparsity pruning strategies: GSDN employs anchor-based strategy with the corresponding anchor-based loss, but in our anchor-free method, we cannot use the anchor-based sparsity pruning. However, the \textit{balanced} FCAF3D with a more lightweight backbone with three feature levels, the voxel size of 0.05m, and the stride of 1 outperforms GSDN in \emph{both accuracy and speed}. Overall, we argue that FCAF3D addresses the 3D object detection in a more efficient way and thus should be preferred.

As can be observed, the all-best results are obtained by the original \textit{accurate} FCAF3D with the HDResNet34 backbone, the voxel size of 0.01m, and the default stride of 2: in this setting, FCAF3D outperforms GSDN by a huge margin (mAP@0.25 of 70.7 against 62.8, mAP@0.5 of 56.0 against 34.8).

Finally, we address the speed issues with the most lightweight HDResNet34:2 backbone having only two feature levels. 
According to the reported values, the \textit{fast} FCAF3D modification with HDResNet34:2 processes 30 scenes per second, while GSDN is able to handle only 20 scenes. While improving the inference speed, we do not sacrifice the superior accuracy: with the voxel size of 0.02m, FCAF3D based on the HDResNet34:2 backbone still outperforms GSDN in both mAP@0.25 and mAP@0.5.

\subsection{Ablation Study}\label{subsec:ablation}

\begin{table*}[ht!]
    \centering \setlength{\tabcolsep}{2.5pt}
    \newcommand{\minitab}[2][l]{\begin{tabular}{#1}#2\end{tabular}}
    \resizebox{1\linewidth}{!}{
    \begin{tabular}{c|c|cccccc}
    \hline
    Ablating & \multirow[l]{2}{*}{Value} & \multicolumn{2}{c}{ScanNet} & \multicolumn{2}{c}{SUN RGB-D} & \multicolumn{2}{c}{S3DIS} \\
    parameter & & mAP@0.25 & mAP@0.5 & mAP@0.25 & mAP@0.5 & mAP@0.25 & mAP@0.5 \\ \hline
    \multirow[l]{3}{*}{\minitab[c]{Voxel \\ size}} & \textbf{0.01} & 71.5 (\textbf{70.7}) & 57.3 (\textbf{56.0}) & 64.2 (\textbf{63.8}) & 48.9 (\textbf{48.2}) & 66.7 (\textbf{64.9}) & 45.9 (\textbf{43.8}) \\ 
    & 0.02 & 66.3 (65.8) & 49.4 (48.6) & 62.3 (62.0) & 46.3 (45.5) & 61.0 (58.5) & 43.8 (38.5) \\
    & 0.03 & 59.6 (59.2) & 42.6 (41.6) & 60.4 (59.7) & 41.6 (41.0) & 55.4 (53.3) & 38.6 (35.0) \\ \hline
    \multirow[l]{3}{*}{\minitab[c]{Number of \\ points}} & 20k & 69.0 (68.1) & 52.8 (52.0) & 63.0 (62.5) & 46.9 (46.5) & 60.1 (58.8) & 45.1 (40.1) \\ 
    & 40k & 67.6 (66.7) & 53.6 (52.2) & 63.4 (63.1) & 47.2 (46.6) & 63.7 (61.2) & 44.8 (42.2) \\
    & \textbf{100k} & 71.5 (\textbf{70.7}) & 57.3 (\textbf{56.0}) & 64.2 (\textbf{63.8}) & 48.9 (\textbf{48.2}) & 66.7 (\textbf{64.9}) & 45.9 (\textbf{43.8}) \\ \hline
    \multirow[l]{2}{*}{Centerness} & No & 71.0 (70.4) & 56.1 (55.1) & 63.8 (63.3) & 48.2 (47.5) & 67.9 (\textbf{65.5}) & 46.0 (43.5) \\ 
    & \textbf{Yes} & 71.5 (\textbf{70.7}) & 57.3 (\textbf{56.0}) & 64.2 (\textbf{63.8}) & 48.9 (\textbf{48.2}) & 66.7 (64.9) & 45.9 (\textbf{43.8}) \\ \hline
    \multirow[l]{3}{*}{\minitab[c]{Center \\ sampling}} & 9 & 70.6 (70.1) & 55.7 (55.0) & 63.8 (63.3) & 48.6 (48.2) & 66.5 (63.6) & 44.4 (42.5) \\ 
    & \textbf{18} & 71.5 (\textbf{70.7}) & 57.3 (\textbf{56.0}) & 64.2 (\textbf{63.8}) & 48.9 (\textbf{48.2}) & 66.7 (\textbf{64.9}) & 45.9 (\textbf{43.8}) \\
    & 27 & 70.2 (69.7) & 55.7 (54.1) & 64.3 (63.8) & 48.7 (47.9) & 65.1 (63.2) & 43.6 (41.7) \\ \hline
    \end{tabular}
    }
    \caption{Results of ablation studies on the voxel size, the number of points (which equals the number of voxels $N_{\text{vox}}$ in pruning), centerness, and center sampling in FCAF3D. The better options are marked bold (actually, these are the default options used to obtain the results in Tab. \ref{tab:results} above). The reported metric value is the best across 25 trials; the average value is given in brackets.}
    \label{tab:ablation_all}
\end{table*}

In this section, we discuss the FCAF3D design choices and investigate how they affect metrics when applied independently in ablation studies. We run experiments with varying voxel size, the number of points in a point cloud $N_\text{pts}$, the number of locations selected by center sampling, and with and without centerness. The results of ablation studies are aggregated in Tab. \ref{tab:ablation_all} for all benchmarks.

\textbf{Voxel size}. Expectedly, with an increasing voxel size, accuracy goes down. We try voxels of 0.03, 0.02, and 0.01 m. We do not experiment with smaller values since inference would take too much time. We attribute the notable gap in mAP between voxel sizes of 0.01 and 0.02 m to the presence of \textit{almost flat} objects, such as doors, pictures, and whiteboards. Namely, with a voxel size of 2 cm, the head would output locations with 16 cm tolerance, but the \textit{almost flat} objects could be less than 16 cm by one of the dimensions. Accordingly, we observe a decrease in accuracy for larger voxel sizes.

\textbf{Number of points}. Similar to 2D images, subsampled point clouds are sometimes referred to as \textit{low-resolution} ones. Accordingly, they contain less information than their \textit{high-resolution} versions. As can be expected, the fewer the points, the lower is detection accuracy. In this series of experiments, we sample 20k, 40k, and 100k points from the entire point cloud, and the obtained metric values revealed a clear dependency between the number of points and mAP. We do not consider larger $N_\text{pts}$ values to be on a par with the existing methods (specifically, GSDN \cite{gwak2020gsdn} uses all points in a point cloud, GroupFree \cite{liu2021group-free} samples 50k points, VoteNet \cite{qi2019votenet} selects 40k points for ScanNet and 20k for SUN RGB-D). We use $N_\text{vox} = N_\text{pts}$ to guide pruning in the neck. When $N_\text{vox}$ exceeds 100k, the inference time increases due to growing sparsity in the neck, while the accuracy improvement is negligible. So we restrict our grid search for $N_\text{pts}$ with 100k and use it as a default value regarding the obtained results.

\textbf{Centerness.} Using centerness improves mAP for the ScanNet and SUN RGB-D datasets. For S3DIS, the results are controversial: the better mAP@0.5 is balanced with a minor decrease of mAP@0.25. Nevertheless, we analyze the results altogether, so we can consider centerness a helpful feature with a small positive effect on the mAP, almost reaching 1\% of mAP@0.5 on ScanNet.

\textbf{Center sampling.} Finally, we study the number of locations selected in center sampling. We select 9 locations, as proposed in FCOS \cite{tian2019fcos}, the entire set of 27 locations, as in ImVoxelNet \cite{rukhovich2021imvoxelnet}, and 18 locations. The latter appeared to be the best choice according to mAP on all the benchmarks.

\subsection{Inference Speed}

Compared to standard convolutions, sparse convolutions are time- and memory-efficient. GSDN authors claim that with sparse convolutions, they process a scene with 78M points covering about 14,000 m\textsuperscript{3} within a single fully convolutional feed-forward pass, using only 5G of GPU memory. FCAF3D uses the same sparse convolutions and the same backbone as GSDN. However, as can be seen in Tab. \ref{tab:ablation_gsdn}, the default FCAF3D is slower than GSDN. This is due to the smaller voxel size: we use 0.01m for a proper multi-level assignment while GSDN uses 0.05m.

To build the fastest method, we use HDResNet34:3 and HDResNet34:2 backbones with only three and two feature levels, respectively. With these modifications, FCAF3D is faster on inference than GSDN (Fig. \ref{fig:fps}). 

For a fair comparison, we re-measure inference speed for GSDN and voting-based methods, as point grouping operation and sparse convolutions have become much faster since the initial release of these methods. In performance tests, we opt for implementations based on the MMdetection3D \cite{2020mmdetection3d} framework to mitigate codebase differences. The reported inference speed for all methods is measured on the same single GPU so they can be directly compared.

\section{Conclusion}

We presented FCAF3D, a first-in-class fully convolutional anchor-free 3D object detection method for indoor scenes. Our method significantly outperforms the previous state-of-the-art on the challenging indoor SUN RGB-D, ScanNet, and S3DIS benchmarks in terms of both mAP and inference speed. We also proposed a novel oriented bounding box parametrization and showed that it improves accuracy for several 3D object detection methods. Moreover, the proposed parametrization allows avoiding any prior assumptions about objects, thus reducing the number of hyperparameters. Overall, FCAF3D with our bounding box parametrization is accurate, scalable, and generalizable at the same time. 

\textbf{Acknowledgment.} We would like to thank Alexey Rukhovich for useful discussions on topology.

\bibliographystyle{splncs04}
\bibliography{egbib}

\definecolor{c0}{RGB}{203,249,0}
\definecolor{c1}{RGB}{205,0,0}
\definecolor{c2}{RGB}{0,52,221}
\definecolor{c3}{RGB}{0,186,0}
\definecolor{c4}{RGB}{255,175,0}
\definecolor{c5}{RGB}{120,0,137}
\definecolor{c6}{RGB}{0,164,187}
\definecolor{c7}{RGB}{41,255,0}
\definecolor{c8}{RGB}{228,0,0}
\definecolor{c9}{RGB}{0,0,187}
\definecolor{c10}{RGB}{0,154,15}
\definecolor{c11}{RGB}{245,222,0}
\definecolor{c12}{RGB}{204,204,204}
\definecolor{c13}{RGB}{0,137,221}
\definecolor{c14}{RGB}{0,224,0}
\definecolor{c15}{RGB}{255,51,0}
\definecolor{c16}{RGB}{105,0,156}
\definecolor{c17}{RGB}{0,170,143}

\appendix

\section{Additional Comments on Mobius Parametrization}

\textbf{Comments on Eq. \ref{eq:q_theta}.}
The OBB heading angle $\theta$ is typically defined as an angle between $x$-axis and a vector towards a center of one of OBB faces. If a frontal face exists, then $\theta$ is defined unambiguously; however, this is not the case for some indoor objects. If a frontal face cannot be chosen unequivocally, there are four possible representations for a single OBB. The heading angle describes a rotation within the $xy$ plane around $z$-axis w.r.t. the OBB center. Therefore, the OBB center $(x,\ y,\ z)$, height $h$, and the OBB size $s=w+l$ are the same for all representations. Meanwhile, the ratio $q=\frac{w}{l}$ of the frontal and lateral OBB faces and the heading angle $\theta$ do vary. Specifically, there are four options for the heading angle: $\theta,\ \theta + \frac{\pi}{2},\ \theta + \pi,\ \theta + \frac{3\pi}{2}$. Swapping frontal and lateral faces gives two ratio options: $q$ and $\frac{1}{q}$. Overall, there are four different tuples $(q, \theta)$ for the same OBB:
\begin{equation*}
    \left(q, \theta \right),\ \left(\frac{1}{q},\ \theta+\frac{\pi}{2} \right),\ \left( q, \theta+\pi \right),\ \left(\frac{1}{q}, \theta+\frac{3\pi}{2} \right).
\end{equation*}

\textbf{Verification of Eq. \ref{eq:embedding}.}
Here, we prove that four different representations of the same OBB from Eq. \ref{eq:q_theta} map to the same point on a Mobius strip by Eq. \ref{eq:embedding}.
\begin{equation*}\small\begin{aligned}
    (q,\ \theta)\ \mapsto\ & \ (\ln(q) \sin(2\theta),\ \ln(q) \cos(2\theta),\ \sin(4\theta),\ \cos(4\theta)) \\
    \left( \frac{1}{q},\ \theta+\frac{\pi}{2} \right) \ \mapsto\ & \ ( \ln ( \frac{1}{q} ) \sin(2\theta + \pi),\  \ln ( \frac{1}{q} ) \cos(2\theta + \pi), \sin(4\theta + 2\pi),\ \cos(4\theta + 2\pi ) \large ) \\
    = & \ (\ln(q) \sin(2\theta),\ \ln(q) \cos(2\theta),\ \sin(4\theta),\ \cos(4\theta)) \\
    (q,\ \theta+\pi)\ \mapsto\ & \ (\ln(q) \sin(2\theta + 2\pi),\ \ln(q) \cos(2\theta + 2\pi), \sin(4\theta + 4\pi),\ \cos(4\theta + 4\pi)) \\
    = & \ (\ln(q) \sin(2\theta),\ \ln(q) \cos(2\theta),\ \sin(4\theta),\ \cos(4\theta)) \\
    \left( \frac{1}{q},\ \theta+\frac{3\pi}{2} \right) \ \mapsto\ & \ ( \ln ( \frac{1}{q} ) \sin(2\theta + 3\pi),\  \ln (\frac{1}{q} ) \cos(2\theta + 3\pi), \sin(4\theta + 6\pi),\ \cos(4\theta + 6\pi) ) \\
    = & \ (\ln(q) \sin(2\theta),\ \ln(q) \cos(2\theta),\ \sin(4\theta),\ \cos(4\theta)) \\
\end{aligned}\end{equation*}

\section{Metric values for Fig. \ref{fig:fps}}

We report inference speed for different methods on ScanNet dataset in Tab. \ref{tab:fps}. The inference speed is measured on the same single NVidia GTX1080Ti.

\begin{table}[h!]
    \centering \setlength{\tabcolsep}{2.5pt}
    \begin{tabular}{l|ccc}
    \hline
    \multirow[l]{2}{*}{Method} & Scenes & \multicolumn{2}{c}{mAP} \\
    & per sec. & 0.25 & 0.5 \\ \hline
    VoteNet\cite{qi2019votenet} & 11.8 & 58.6 & 33.5 \\
    GSDN\cite{gwak2020gsdn} & 20.1 & 62.8 & 34.8 \\
    H3DNet\cite{zhang2020h3dnet} & 4.9 & 67.2 & 48.1 \\
    BRNet\cite{cheng2021brnet} & 10.3 & 66.1 & 50.9 \\
    3DETR\cite{misra20213detr} & 3.1 & 62.7 & 37.5 \\
    3DETR-m\cite{misra20213detr} & 3.1 & 65.0 & 47.0 \\
    GroupFree\cite{liu2021group-free} & 6.6 & 69.1 & 52.8 \\
    FCAF3D & 8.0 & \textbf{71.5} & \textbf{57.3} \\
    \ \  w/ 3 levels & 12.2 & 69.8 & 53.6 \\
    \ \  w/ 2 levels & \textbf{31.5} & 63.1 & 46.8 \\ \hline
    \end{tabular}
    \caption{Results of 3D object detection methods that accept point clouds on ScanNet.}
    \label{tab:fps}
\end{table}

\section{Per-category results}

\textbf{ScanNet.} Tab. \ref{tab:scannet25} contains per-category AP@0.25 scores for 18 object categories for the ScanNet dataset. For 12 out of 18 categories, FCAF3D outperforms other methods. The largest quality gap can be observed for \textit{window} (60.2 against 53.7), \textit{picture} (29.9 against 18.6), and \textit{other furniture} (65.4 against 56.4) categories.

\begin{table}[h!]
    \centering \small
    \begingroup 
    \resizebox{1\linewidth}{!}{
    \begin{tabular}{l|cccccccccccccccccc|c}
    \hline
    Method & cab & bed & chair & sofa & tabl & door & wind & bkshf & pic & cntr & desk & curt & fridg & showr & toil & sink & bath & ofurn & mAP \\ \hline
    VoteNet\cite{qi2019votenet} & 36.3 & 87.9 & 88.7 & 89.6 & 58.8 & 47.3 & 38.1 & 44.6 & \phantom{0}7.8 & 56.1 & 71.7 & 47.2 & 45.4 & 57.1 & 94.9 & 54.7 & 92.1 & 37.2 & 58.7 \\
    GSDN\cite{gwak2020gsdn} & 41.6 & 82.5 & 92.1 & 87.0 & 61.1 & 42.4 & 40.7 & 51.5 & 10.2 & 64.2 & 71.1 & 54.9 & 40.0 & 70.5 & \textbf{100} & 75.5 & 93.2 & 53.1 & 62.8 \\
    H3DNet\cite{zhang2020h3dnet} & 49.4 & 88.6 & 91.8 & 90.2 & 64.9 & 61.0 & 51.9 & 54.9 & 18.6 & 62.0 & 75.9 & 57.3 & 57.2 & 75.3 & 97.9 & 67.4 & 92.5 & 53.6 & 67.2 \\ 
    GroupFree\cite{liu2021group-free} & 52.1 & \textbf{92.9} & 93.6 & 88.0 & \textbf{70.7} & 60.7 & 53.7 & 62.4 & 16.1 & 58.5 & \textbf{80.9} & \textbf{67.9} & 47.0 & 76.3 & 99.6 & 72.0 & \textbf{95.3} & 56.4 & 69.1 \\
    FCAF3D & \textbf{57.2} & 87.0 & \textbf{95.0} & \textbf{92.3} & 70.3 & \textbf{61.1} & \textbf{60.2} & \textbf{64.5} & \textbf{29.9} & \textbf{64.3} & 71.5 & 60.1 & \textbf{52.4} & \textbf{83.9} & 99.9 & \textbf{84.7} & 86.6 & \textbf{65.4} & \textbf{71.5} \\ \hline
    \end{tabular}
    }\endgroup
    \caption{Per-category AP@0.25 scores for 18 object categories from the ScanNet dataset.}
    \label{tab:scannet25}
\end{table}

Tab. \ref{tab:scannet50} shows per-category AP@0.5 scores. According to the reported values, FCAF3D is the best at detecting objects of 13 out of 18 categories. The most significant improvement is achieved for \textit{cabinet} (35.8 against 26.0), \textit{sofa} (85.2 against 70.7), \textit{picture} (17.9 against 7.8), shower (64.2 against 44.1), and \textit{sink} (52.6 against 37.4).

\begin{table}[h!]
    \centering \small
    \begingroup 
    \resizebox{1\linewidth}{!}{
    \begin{tabular}{l|cccccccccccccccccc|c}
    \hline
    Method & cab & bed & chair & sofa & tabl & door & wind & bkshf & pic & cntr & desk & curt & fridg & showr & toil & sink & bath & ofurn & mAP \\ \hline
    VoteNet\cite{qi2019votenet} & \phantom{0}8.1 & 76.1 & 67.2 & 68.8 & 42.4 & 15.3 & \phantom{0}6.4 & 28.0 & \phantom{0}1.3 & \phantom{0}9.5 & 37.5 & 11.6 & 27.8 & 10.0 & 86.5 & 16.8 & 78.9 & 11.7 & 33.5 \\
    GSDN\cite{gwak2020gsdn} & 13.2 & 74.9 & 75.8 & 60.3 & 39.5 & \phantom{0}8.5 & 11.6 & 27.6 & \phantom{0}1.5 & \phantom{0}3.2 & 37.5 & 14.1 & 25.9 & \phantom{0}1.4 & 87.0 & 37.5 & 76.9 & 30.5 & 34.8 \\
    H3DNet\cite{zhang2020h3dnet} & 20.5 & 79.7 & 80.1 & 79.6 & 56.2 & 29.0 & 21.3 & 45.5 & \phantom{0}4.2 & 33.5 & 50.6 & 37.3 & 41.4 & 37.0 & 89.1 & 35.1 & 90.2 & 35.4 & 48.1 \\ 
    GroupFree\cite{liu2021group-free} & 26.0 & 81.3 & 82.9 & 70.7 & \textbf{62.2} & 41.7 & 26.5 & 55.8 & \phantom{0}7.8 & \textbf{34.7} & \textbf{67.2} & 43.9 & 44.3 & 44.1 & \textbf{92.8} & 37.4 & \textbf{89.7} & 40.6 & 52.8 \\
    FCAF3D & \textbf{35.8} & \textbf{81.5} & \textbf{89.8} & \textbf{85.0} & 62.0 & \textbf{44.1} & \textbf{30.7} & \textbf{58.4} & \textbf{17.9} & 31.3 & 53.4 & \textbf{44.2} & \textbf{46.8} & \textbf{64.2} & 91.6 & \textbf{52.6} & 84.5 & \textbf{57.1} & \textbf{57.3} \\ \hline
    \end{tabular}} \endgroup
    \caption{AP@0.5 scores for 18 object categories from the ScanNet dataset.}
    \label{tab:scannet50}
\end{table}

\textbf{SUN RGB-D.} Per-category AP@0.25 scores for the 10 most common object categories for the SUN RGB-D benchmark are reported in Tab. \ref{tab:sunrgbd25}. Compared to other methods, FCAF3D is more accurate at detecting objects of 7 out of 10 categories. In this experiment, the quality gap is not so dramatic: it equals 4.1 \% for \textit{desk} and 5.2 \% for \textit{night stand}; for the rest categories, it does not exceed 2 \%. FCAF3D achieves a 1.2 \% better mAP@0.25 compared to the closest competitor GroupFree.

\begin{table}[!h]
    \centering \small
    \begingroup \setlength{\tabcolsep}{2.5pt}
    \begin{tabular}{l|cccccccccc|c}
    \hline
    Method & bath & bed & bkshf & chair & desk & dresser & nstand & sofa & table & toilet & mAP \\ \hline
    VoteNet\cite{qi2019votenet} & 74.4 & 83.0 & 28.8 & 75.3 & 22.0 & 29.8 & 62.2 & 64.0 & 47.3 & 90.1 & 57.7 \\
    H3DNet\cite{zhang2020h3dnet} & 73.8 & 85.6 & 31.0 & 76.7 & 29.6 & 33.4 & 65.5 & 66.5 & 50.8 & 88.2 & 60.1 \\
    GroupFree\cite{liu2021group-free} & \textbf{80.0} & 87.8 & 32.5 & 79.4 & 32.6 & 36.0 & 66.7 & \textbf{70.0} & \textbf{53.8} & 91.1 & 63.0 \\
    FCAF3D & 79.0 & \textbf{88.3} & \textbf{33.0} & \textbf{81.1} & \textbf{34.0} & \textbf{40.1} & \textbf{71.9} & 69.7 & 53.0 & \textbf{91.3} & \textbf{64.2} \\ \hline
    \end{tabular} \endgroup
    \caption{AP@0.25 scores for 10 object categories from the SUN RGB-D dataset.}
    \label{tab:sunrgbd25}
\end{table}

For SUN RGB-D, the superiority of the proposed method is more noticeable when analyzing on per-category AP@0.5. As shown in Tab. \ref{tab:sunrgbd50}, FCAF3D outperforms the competitors for 9 out of 10 object categories. For some categories, there is a significant margin: e.g., 30.1 against 21.9 for \textit{dresser}, 59.8 against 49.8 for \textit{night stand}, and 35.5 against 29.2 for \textit{table}. Respectively, FCAF3D surpasses other methods by more than 3.5 \% in terms of mAP@0.5. 

\begin{table}[!h]
    \centering \small
    \begingroup \setlength{\tabcolsep}{2.5pt}
    \resizebox{1\linewidth}{!}{
    \begin{tabular}{l|cccccccccc|c}
    \hline
    Method & bath & bed & bkshf & chair & desk & dresser & nstand & sofa & table & toilet & mAP \\ \hline
    H3DNet\cite{zhang2020h3dnet} & 47.6 & 52.9 & 8.6 & 60.1 & \phantom{0}8.4 & 20.6 & 45.6 & 50.4 & 27.1 & 69.1 & 39.0 \\
    GroupFree\cite{liu2021group-free} & 64.0 & 67.1 & \textbf{12.4} & 62.6 & 14.5 & 21.9 & 49.8 & \textbf{58.2} & 29.2 & 72.2 & 45.2 \\
    FCAF3D & \textbf{66.2} & \textbf{69.8} & 11.6 & \textbf{68.8} & \textbf{14.8} & \textbf{30.1} & \textbf{59.8} & \textbf{58.2} & \textbf{35.5} & \textbf{74.5} & \textbf{48.9} \\ \hline
    \end{tabular} 
    } \endgroup
    \caption{AP@0.5 scores for 10 object categories from the SUN RGB-D dataset.}
    \label{tab:sunrgbd50}
\end{table}

\textbf{S3DIS.} The results of the proposed method in comparison with GSDN are presented in Tab. \ref{tab:s3dis25} and Tab. \ref{tab:s3dis50}. In terms of AP@0.25, FCAF3D is far more accurate when detecting \textit{sofas}, \textit{bookcases}, and \textit{whiteboards}. Most notably, FCAF3D achieves an impressive AP@0.25 of 92.4 for the \textit{sofa} category, leaving GSDN with AP@0.25 of 20.8 far behind. The difference in mAP in favor of the proposed method is almost 19 \%.

\begin{table}[!h]
    \centering \small
    \begingroup \setlength{\tabcolsep}{2.5pt}
    \begin{tabular}{l|ccccc|c}
    \hline
    Method & table & chair & sofa & bkcase & board & mAP \\ \hline
    GSDN\cite{gwak2020gsdn} & \textbf{73.7} & \textbf{98.1} & 20.8 & 33.4 & 12.9 & 47.8 \\
    FCAF3D & 69.7 & 97.4 & \textbf{92.4} & \textbf{36.7} & \textbf{37.3} & \textbf{66.7} \\ \hline
    \end{tabular} \endgroup
    \caption{Per-category AP@0.25 scores for 5 object categories from the S3DIS dataset.}
    \label{tab:s3dis25}
\end{table}

In terms of AP@0.5, FCAF3D outperforms GSDN by a large margin for each category. Similar to AP@0.25, the accuracy gap for the \textit{sofa} category is the most dramatic: with an AP@0.25 of 70.1, FCAF3D is an order of magnitude more accurate than GSDN, which has only 6.1. Accordingly, FCAF3D has an approximately 1.8 times larger mAP compared to GSDN.

\begin{table}[!h]
    \centering \small
    \begingroup \setlength{\tabcolsep}{2.5pt}
    \begin{tabular}{l|ccccc|c}
    \hline
    Method & table & chair & sofa & bkcase & board & mAP \\ \hline
    GSDN\cite{gwak2020gsdn} & 36.6 & 75.3 & \phantom{0}6.1 & \phantom{0}6.5 & \phantom{0}1.2 & 25.1 \\
    FCAF3D & \textbf{45.4} & \textbf{88.3} & \textbf{70.1} & \textbf{19.5} & \textbf{\phantom{0}5.6} & \textbf{45.9} \\ \hline
    \end{tabular} \endgroup
    \caption{AP@0.5 scores for 5 object categories from the S3DIS dataset.}
    \label{tab:s3dis50}
\end{table}

\clearpage

\section{Visualization}

This section contains additional visualizations of the results of 3D object detection for all three benchmarks. The ground truth and estimated 3D object bounding boxes are drawn over the corresponding point clouds. Objects of different categories are marked with different colors.

\begin{figure}[h!]
\centering
\setlength{\tabcolsep}{2pt}
\begin{tabular}{cc}
    \includegraphics[width=0.45\linewidth]{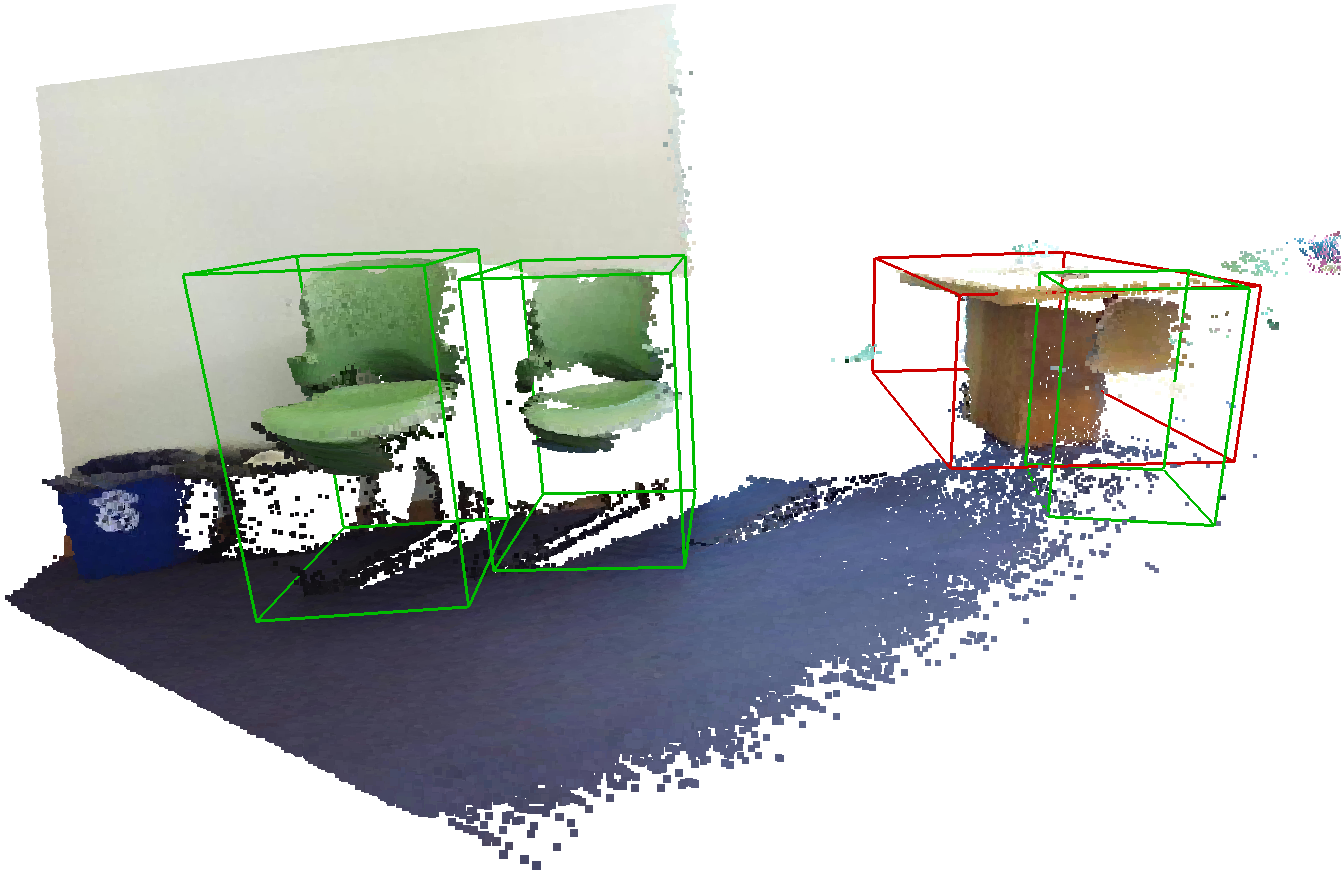} &
    \includegraphics[width=0.45\linewidth]{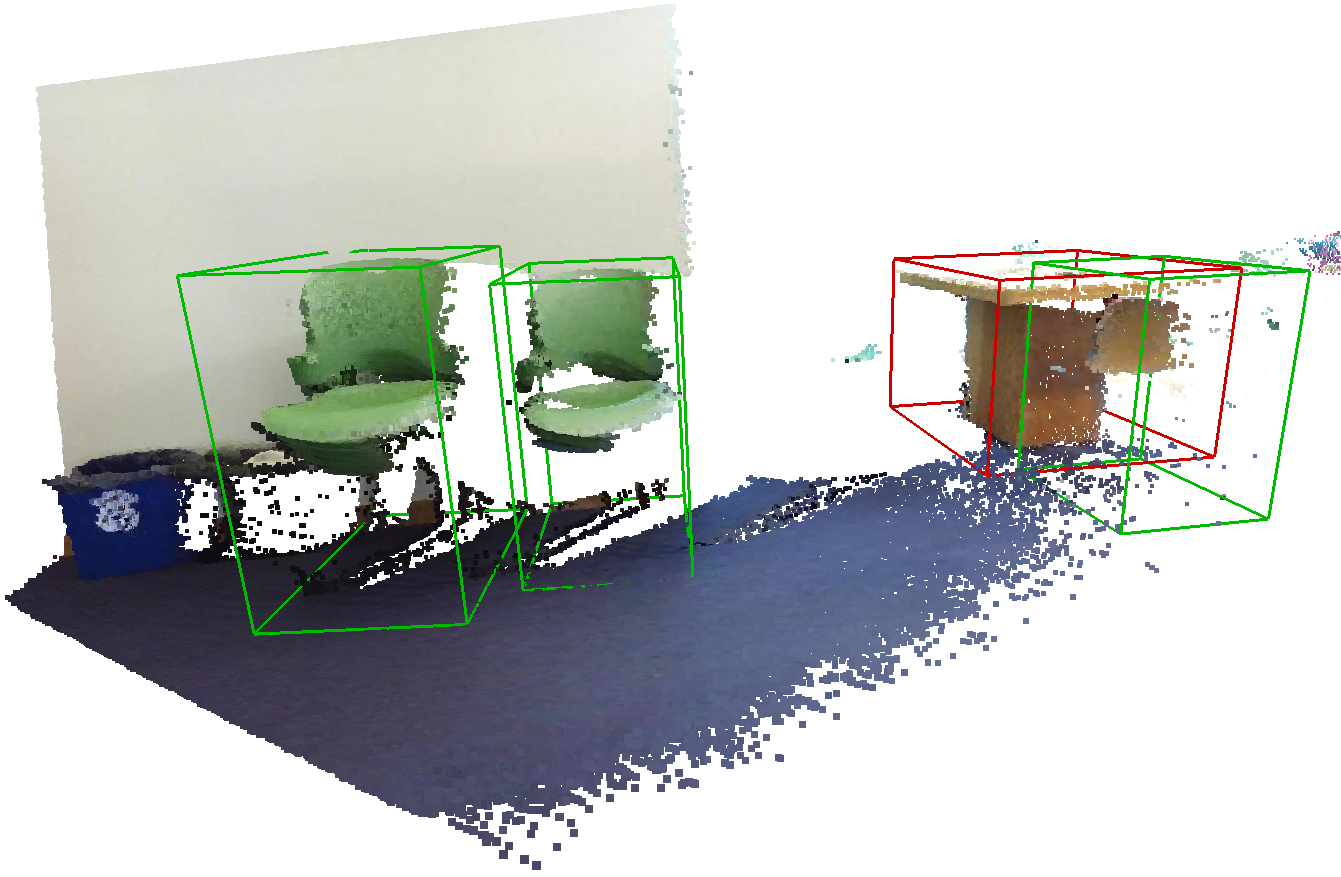} \\
    \includegraphics[width=0.45\linewidth]{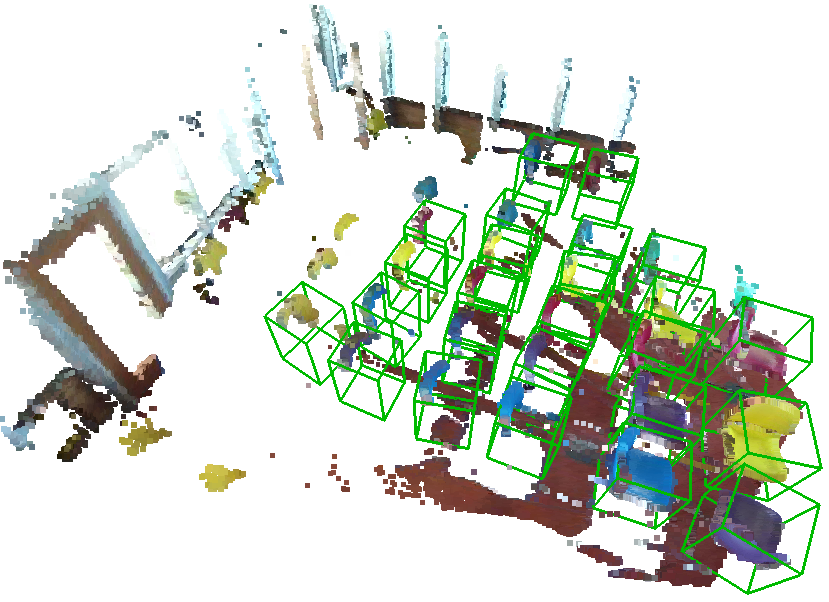} &
    \includegraphics[width=0.45\linewidth]{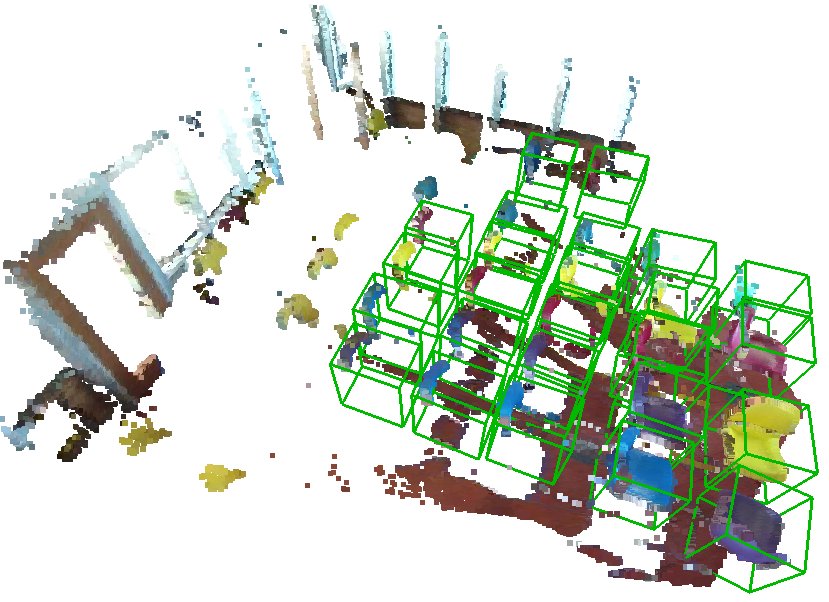} \\
    \includegraphics[width=0.45\linewidth]{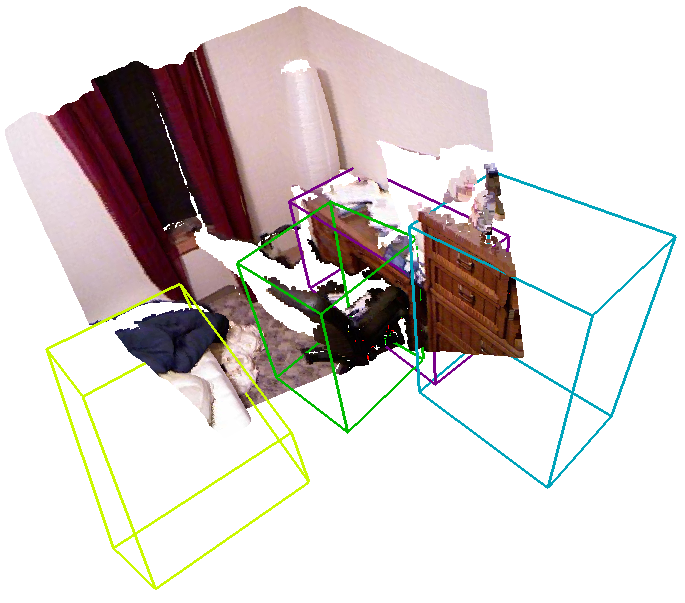} &
    \includegraphics[width=0.45\linewidth]{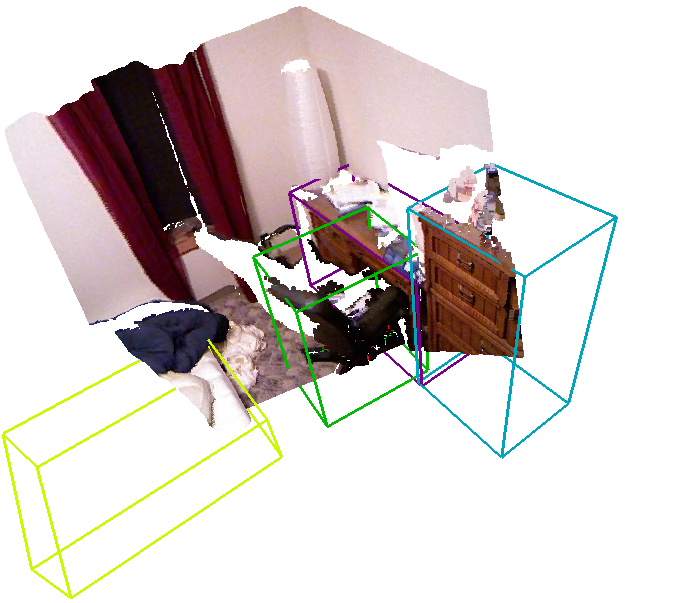}
\end{tabular}
\caption{The point cloud from SUN RGB-D with OBBs. The color of a bounding box denotes the object category: \textbf{\textcolor{c0}{bed}}, \textbf{\textcolor{c3}{chair}}, \textbf{\textcolor{c5}{desk}}, \textbf{\textcolor{c6}{dresser}}, \textbf{\textcolor{c1}{table}} (only categories that are present in the pictures are listed). Left: estimated with FCAF3D, right: ground truth.}
\label{fig:sunrgbd_more_examples}
\end{figure}

\begin{figure}[h!]
\centering
\setlength{\tabcolsep}{2pt}
\begin{tabular}{cc}
    \includegraphics[width=0.45\linewidth]{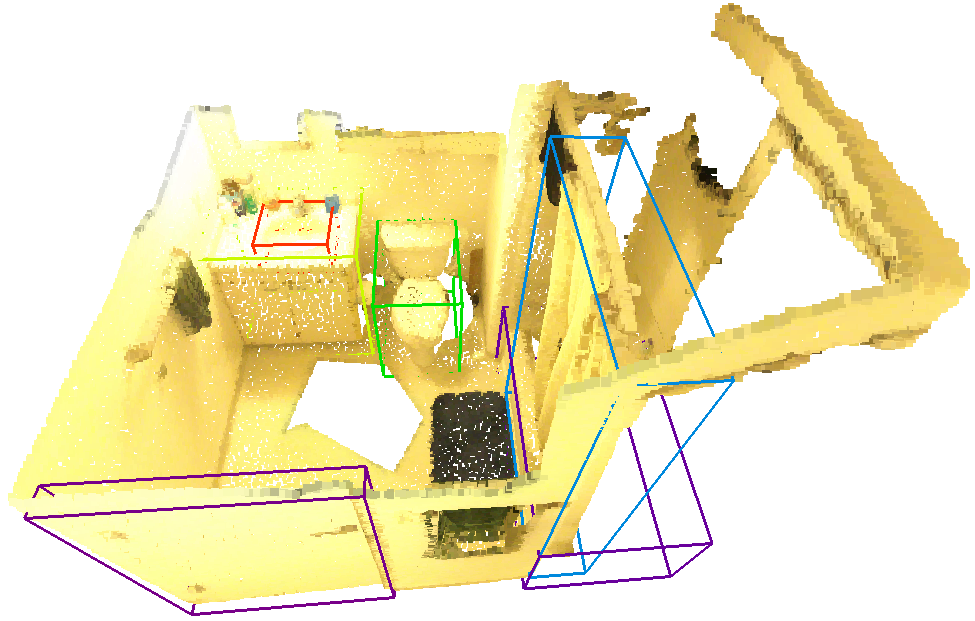} &
    \includegraphics[width=0.45\linewidth]{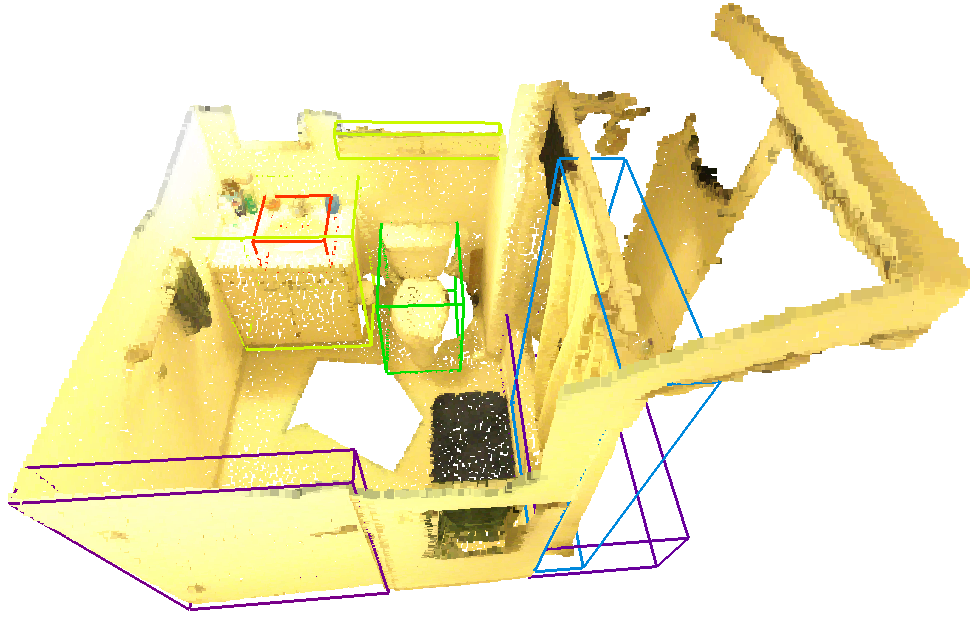} \\ 
    \includegraphics[width=0.45\linewidth]{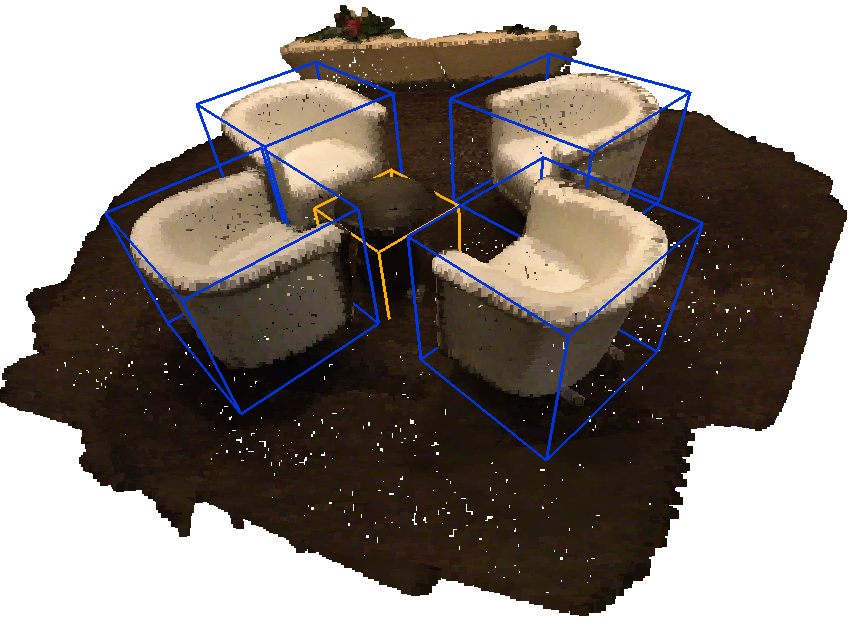} &
    \includegraphics[width=0.45\linewidth]{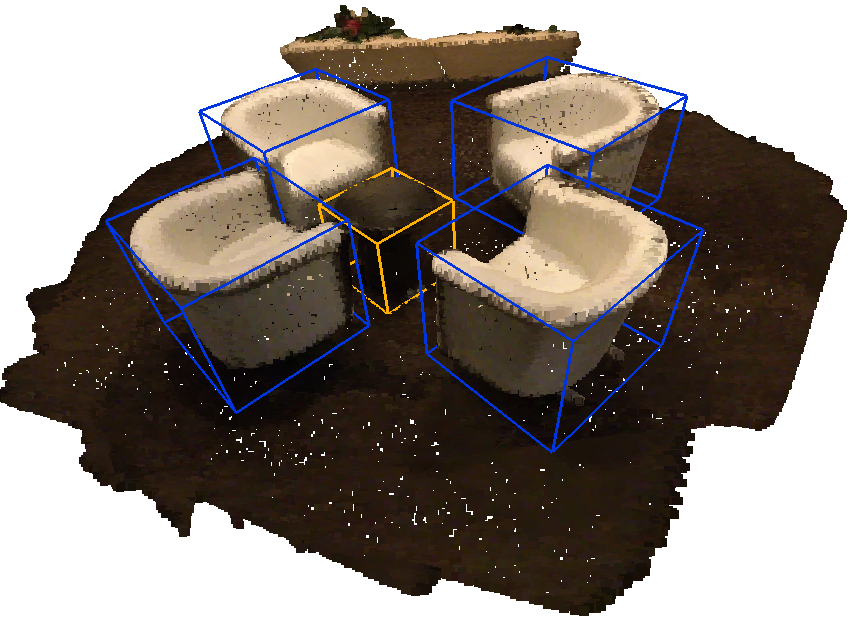} \\
    \includegraphics[width=0.45\linewidth]{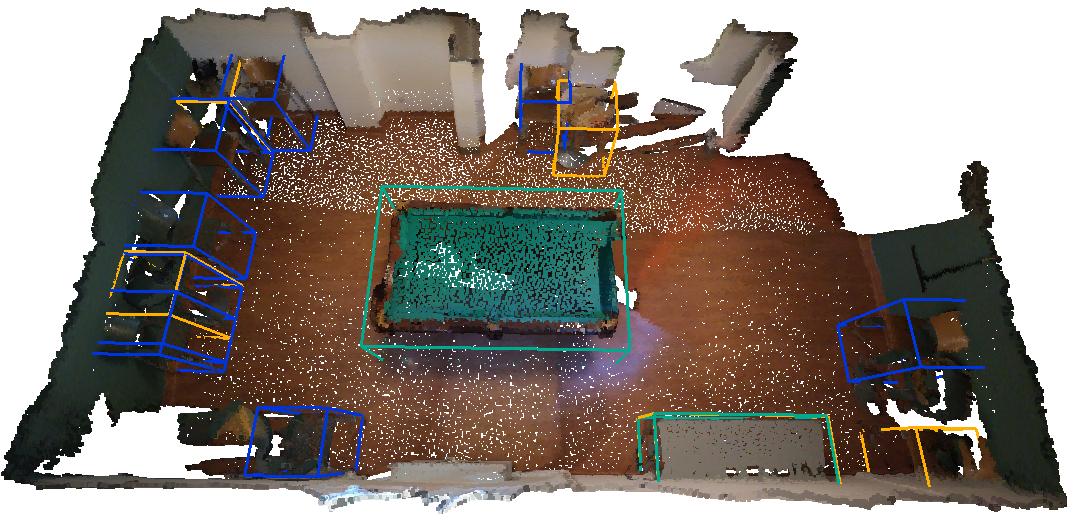} &
    \includegraphics[width=0.45\linewidth]{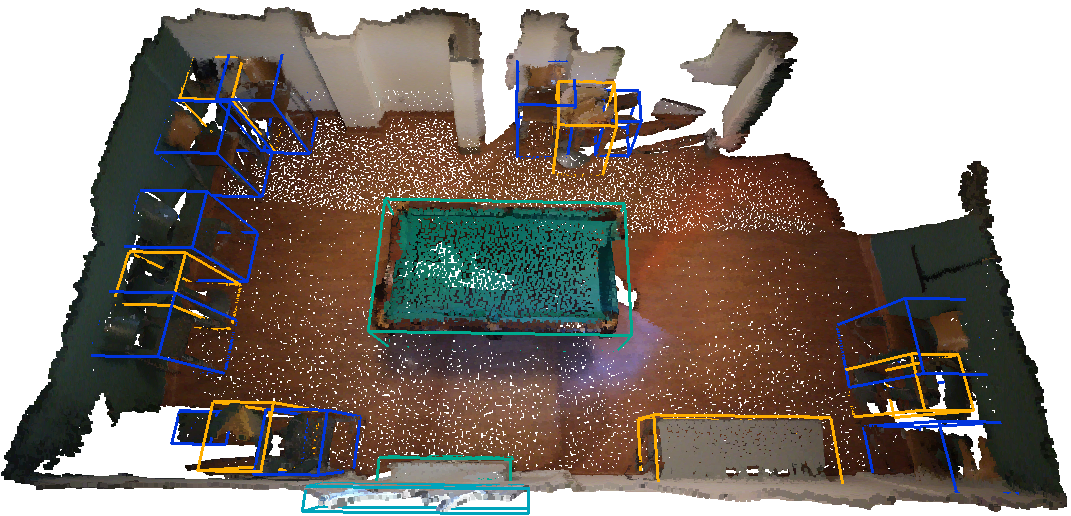} \\
    \includegraphics[width=0.45\linewidth]{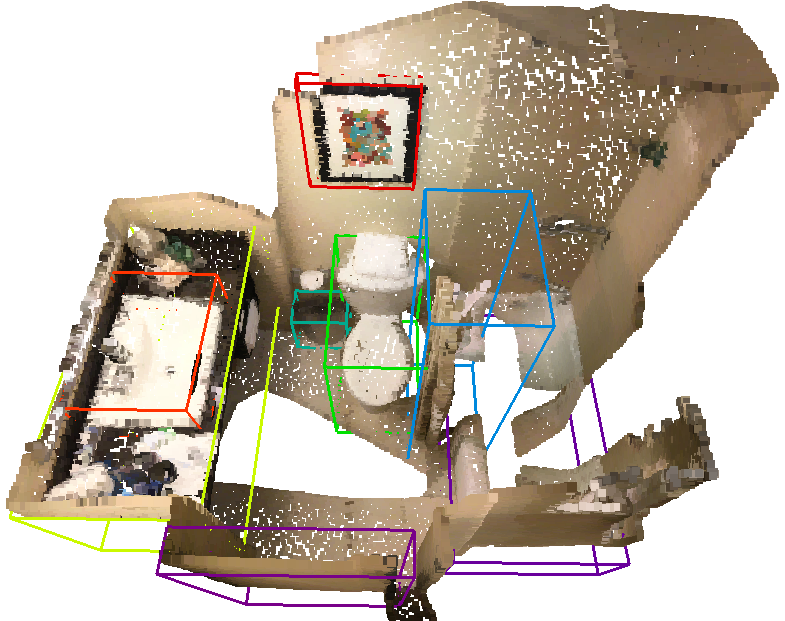} &
    \includegraphics[width=0.45\linewidth]{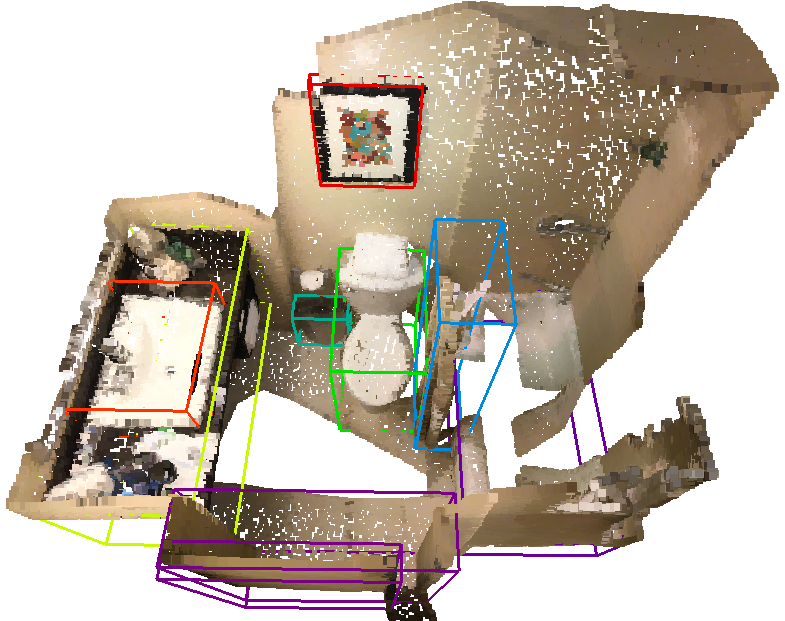}
\end{tabular}
\caption{The point cloud from ScanNet with AABBs. The color of a bounding box denotes the object category: \textbf{\textcolor{c0}{cabinet}}, \textbf{\textcolor{c2}{chair}}, \textbf{\textcolor{c3}{sofa}}, \textbf{\textcolor{c4}{table}}, \textbf{\textcolor{c5}{door}}, \textbf{\textcolor{c6}{window}}, \textbf{\textcolor{c7}{bookshelf}}, \textbf{\textcolor{c8}{picture}}, \textbf{\textcolor{c9}{counter}}, \textbf{\textcolor{c10}{desk}}, \textbf{\textcolor{c13}{shower curtain}}, \textbf{\textcolor{c14}{toilet}}, \textbf{\textcolor{c15}{sink}}, \textbf{\textcolor{c16}{bathtub}}, \textbf{\textcolor{c17}{other furniture}} (only categories that are present in the pictures are listed). Left: estimated with FCAF3D, right: ground truth.}
\label{fig:scannet_more_examples}
\end{figure}

\begin{figure}[h!]
\centering
\setlength{\tabcolsep}{2pt}
\begin{tabular}{cc}
    \includegraphics[width=0.45\linewidth]{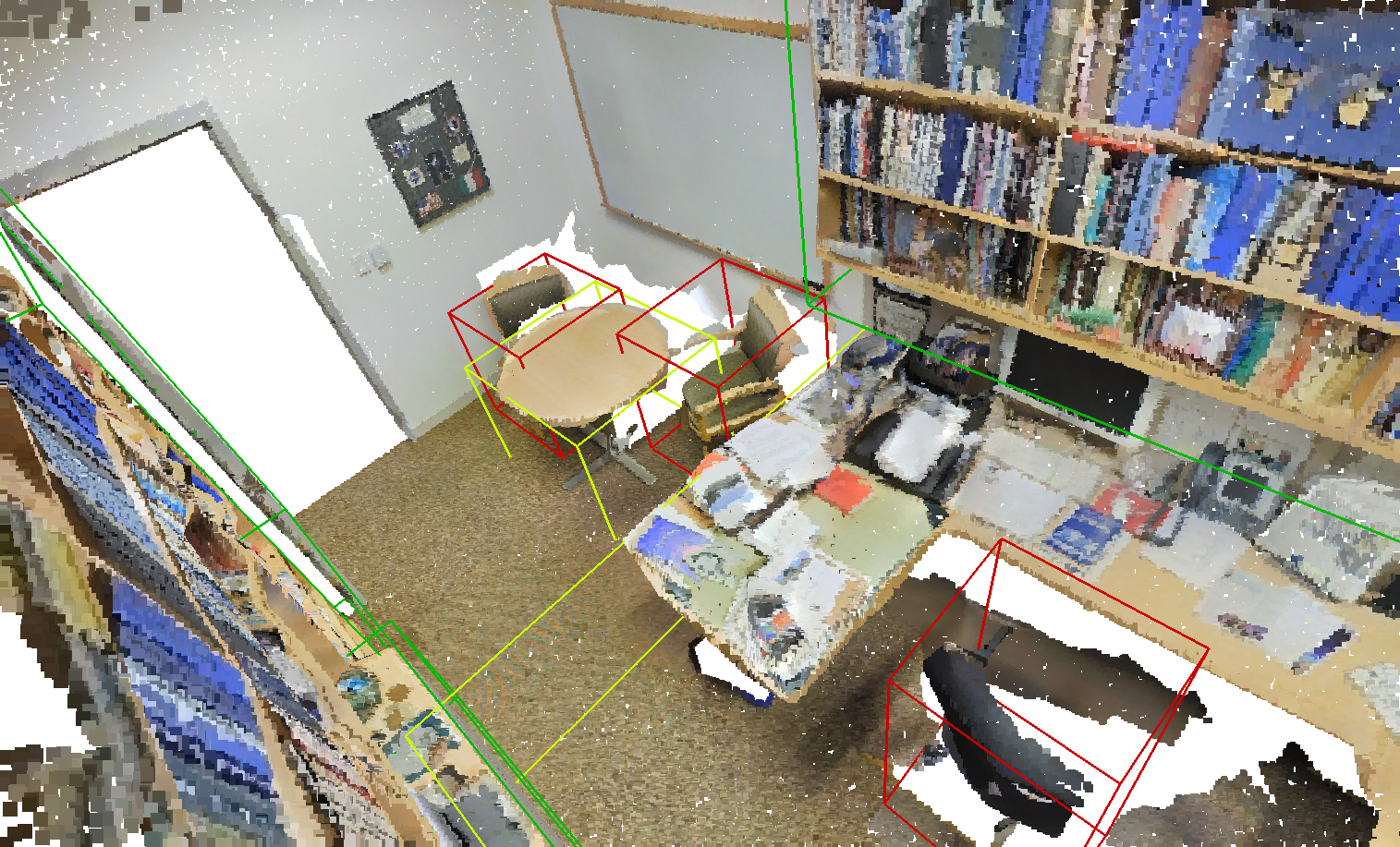} &
    \includegraphics[width=0.45\linewidth]{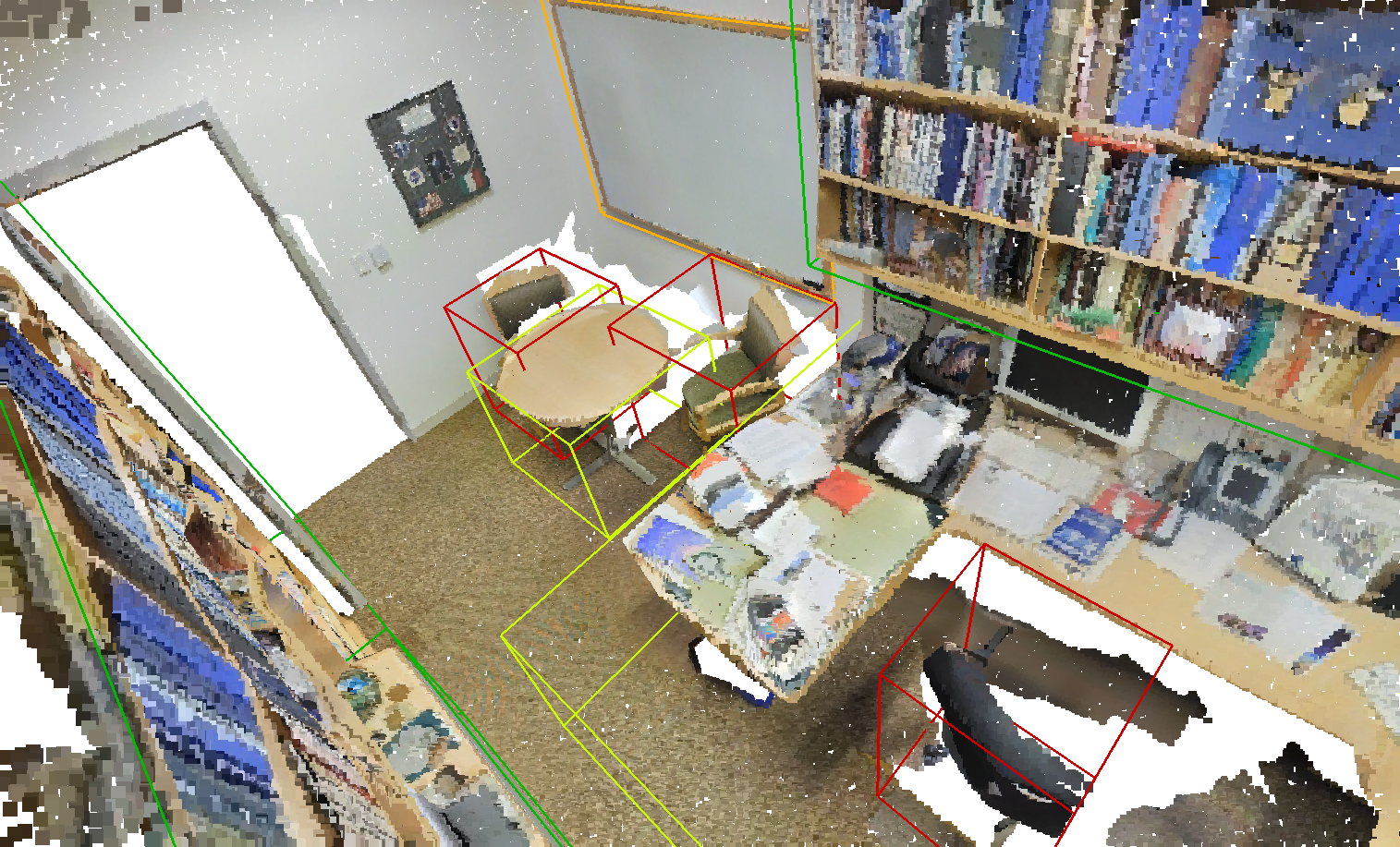} \\
    \includegraphics[width=0.45\linewidth]{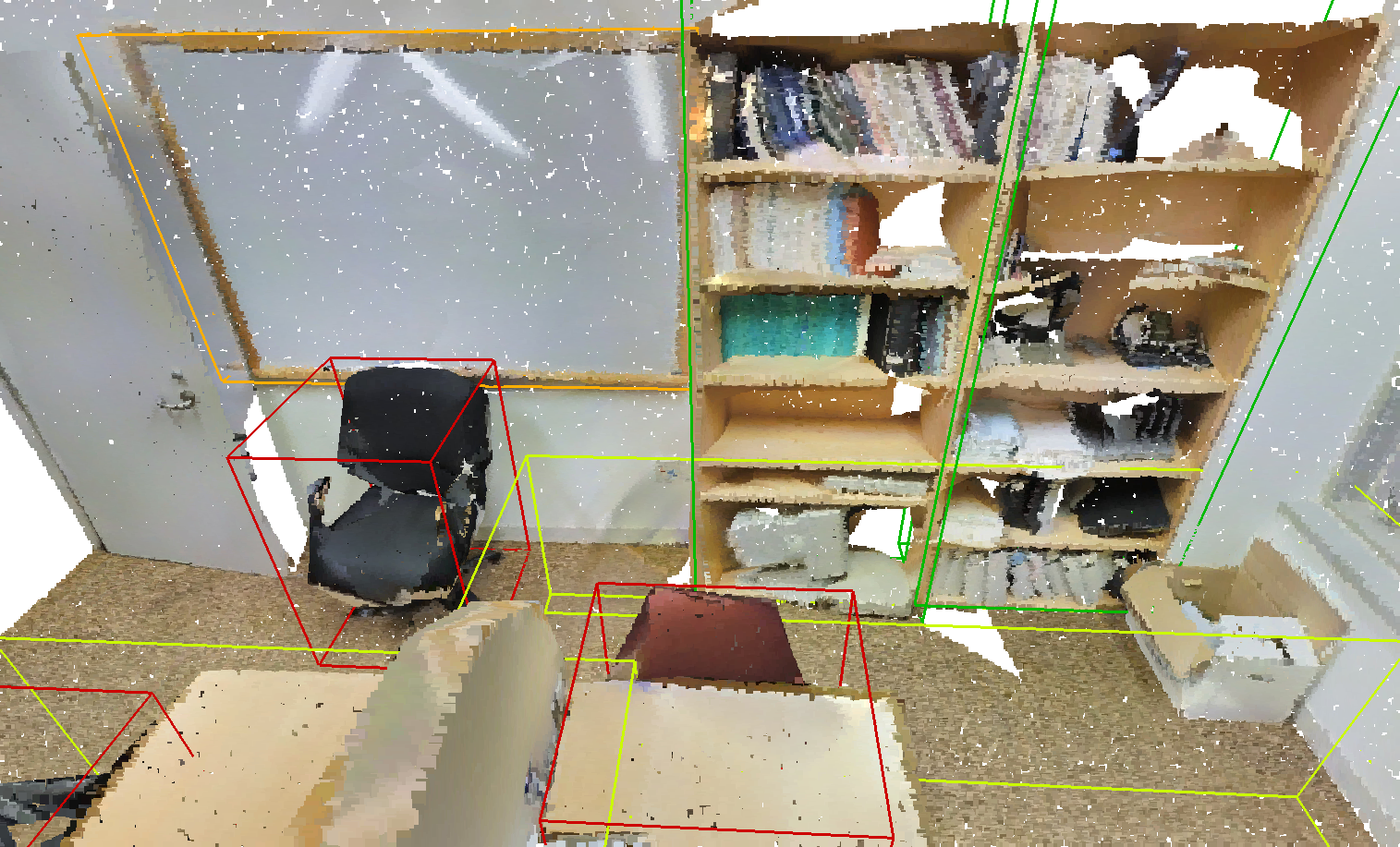} &
    \includegraphics[width=0.45\linewidth]{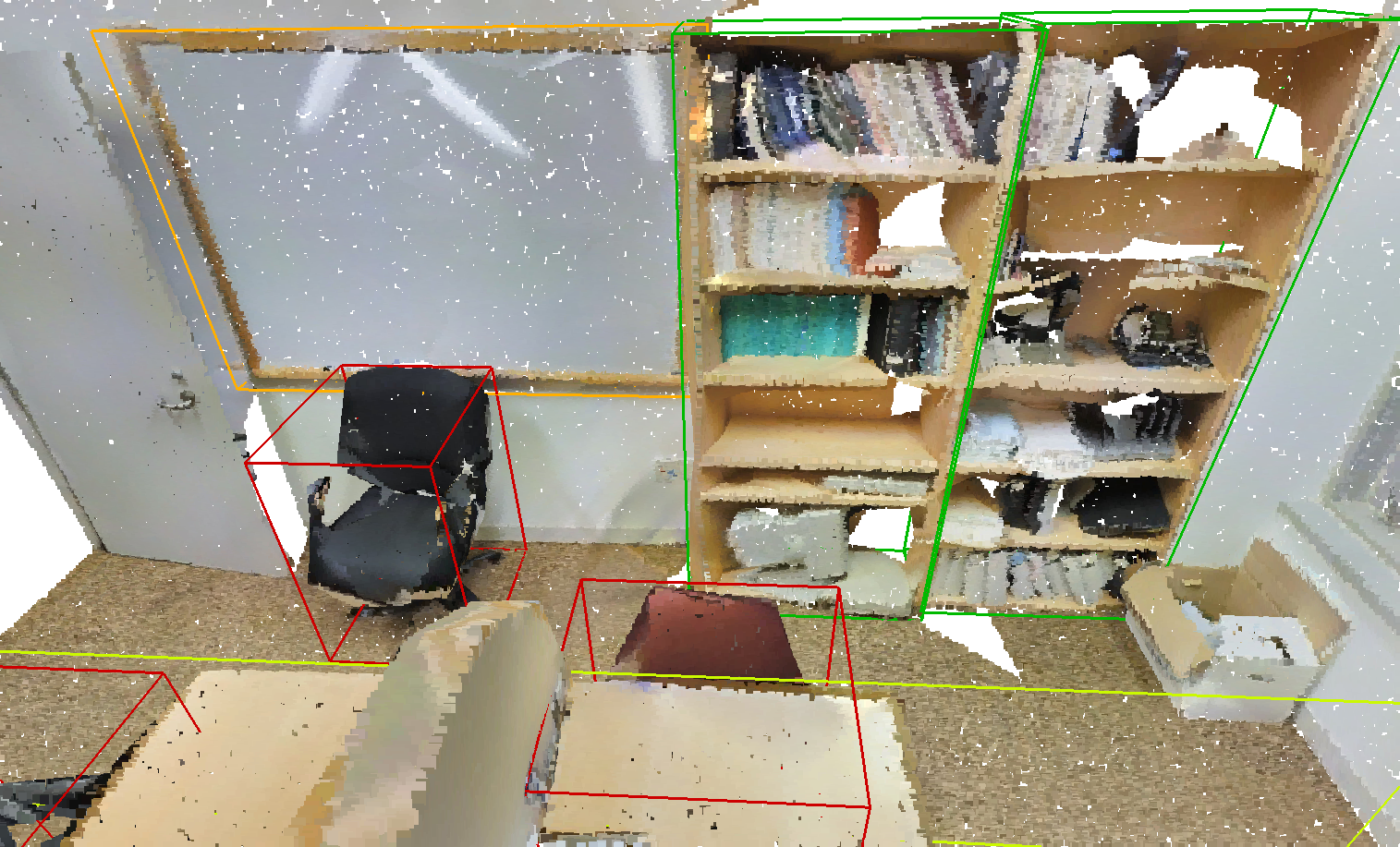} \\
    \includegraphics[width=0.45\linewidth]{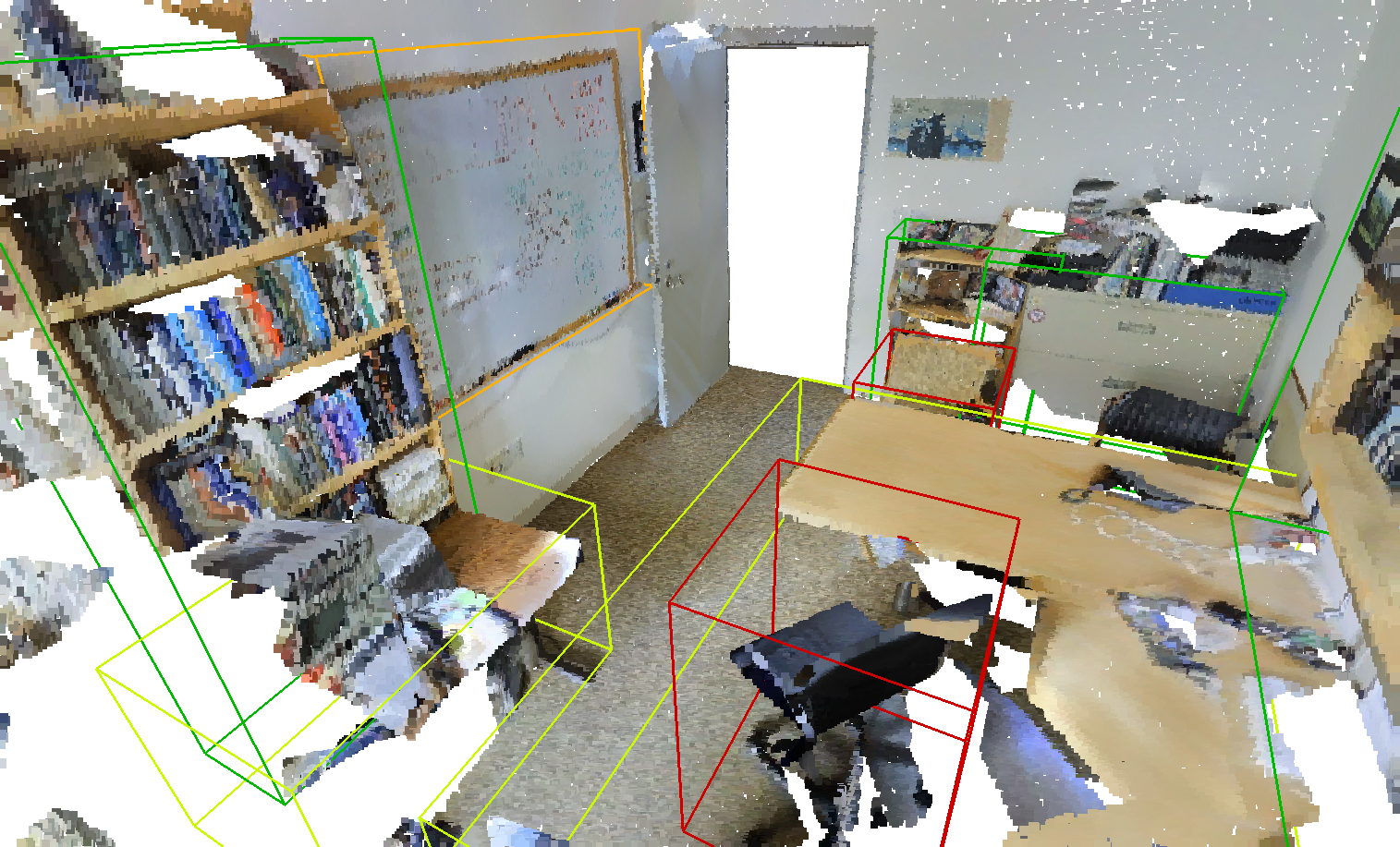} &
    \includegraphics[width=0.45\linewidth]{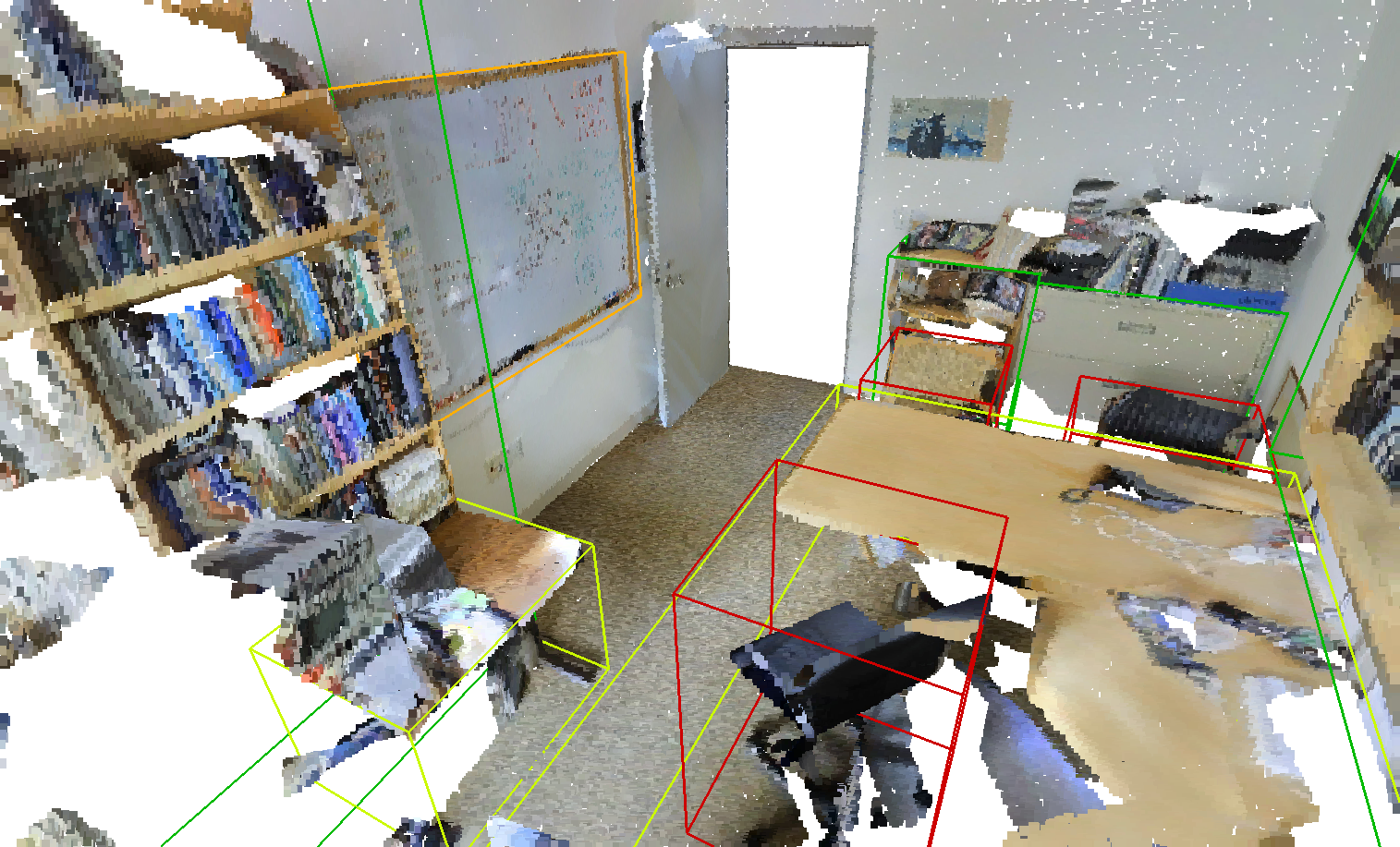} \\
    \includegraphics[width=0.45\linewidth]{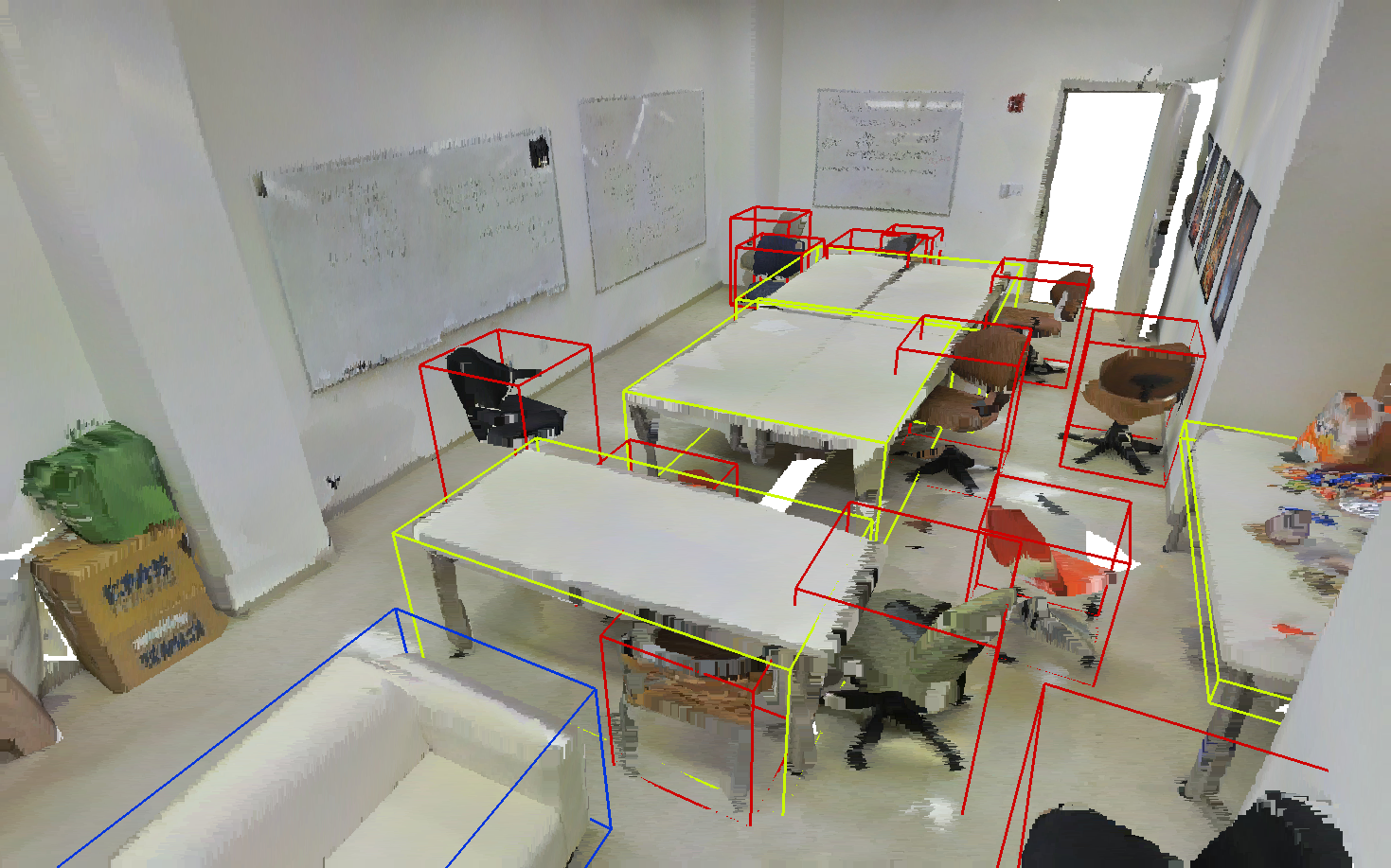} &
    \includegraphics[width=0.45\linewidth]{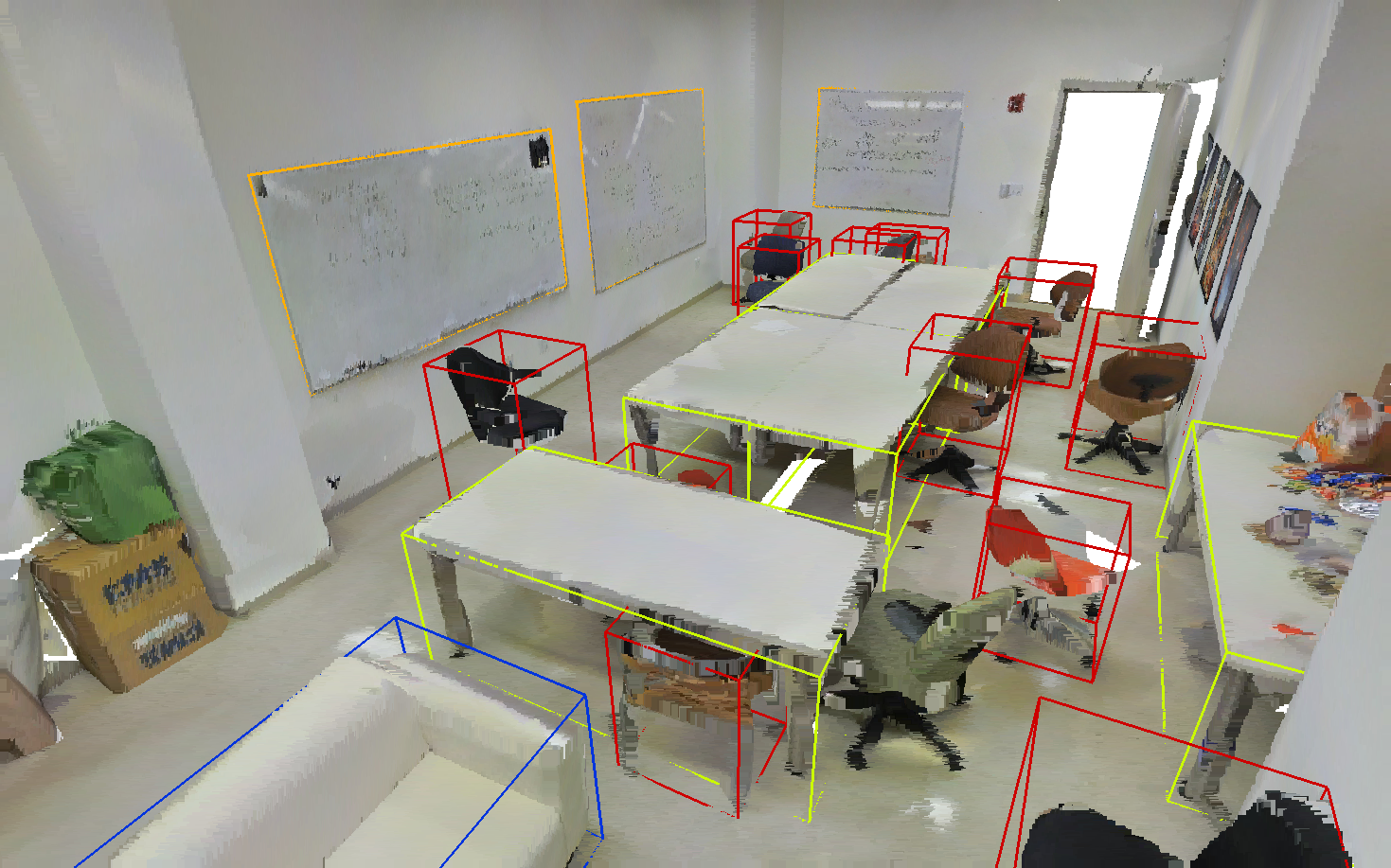}
\end{tabular}
\caption{The point cloud from S3DIS with AABBs. The color of a bounding box denotes the object category: \textbf{\textcolor{c0}{table}}, \textbf{\textcolor{c1}{chair}}, \textbf{\textcolor{c2}{sofa}}, \textbf{\textcolor{c3}{bookcase}}, \textbf{\textcolor{c4}{whiteboard}}. Left: estimated with FCAF3D, right: ground truth.}
\label{fig:s3dis_more_examples}
\end{figure}

\end{document}